\documentclass[10pt,journal,cspaper,compsoc]{IEEEtran}

\usepackage{booktabs,caption}
\usepackage{graphicx}
\usepackage{mathtools}
\usepackage{enumerate}
\usepackage {dsfont}
\usepackage{times}
\usepackage{epsfig}
\usepackage{balance}
\usepackage{textcomp}
\usepackage{color}
\usepackage{amsmath,amssymb} 
\usepackage{cite}
\usepackage{multirow}
\usepackage{url}
\usepackage{mathabx}

\ifCLASSOPTIONcompsoc
\else
\fi
\ifCLASSINFOpdf
\else
\fi
\begin{document}
%
\title{Learning a Deep Model for Human Action Recognition from Novel Viewpoints}

\author{Hossein~Rahmani, Ajmal~Mian and Mubarak Shah
\IEEEcompsocitemizethanks{\IEEEcompsocthanksitem H. Rahmani and A. Mian are with the School of Computer Science and Software Engineering, The University of Western Australia, 35 Stirling Highway,Crawley, Western Australia, 6009.\protect\\
E-mail: hossein@csse.uwa.edu.au, ajmal.mian@uwa.edu.au
}
\IEEEcompsocitemizethanks{\IEEEcompsocthanksitem M. Shah is with with the School of Electric Engineering and Computer Science, University of Central Florida, Orlando, FL 32816-2365 USA.\protect\\
E-mail: shah@eecs.ucf.edu
}
\thanks{}}

\markboth{Manuscript for Review, 2015}%
{Shell \MakeLowercase{\textit{et al.}}: Bare Demo of IEEEtran.cls for Computer Society Journals}

\IEEEcompsoctitleabstractindextext{%
\begin{abstract}
Recognizing human actions from unknown and unseen (novel) views is a challenging problem. We propose a Robust Non-Linear Knowledge Transfer Model (R-NKTM) for human action recognition from novel views. The proposed R-NKTM is a deep fully-connected neural network that transfers knowledge of human actions from any unknown view to a shared high-level virtual view by finding a non-linear virtual path that connects the views. The R-NKTM is learned from dense trajectories of synthetic 3D human models fitted to real motion capture data and generalizes to real videos of human actions. The strength of our technique is that we learn a single R-NKTM for all actions and all viewpoints for knowledge transfer of any real human action video without the need for re-training or fine-tuning the model. Thus, R-NKTM can efficiently scale to incorporate new action classes. R-NKTM is learned with dummy labels and does not require knowledge of the camera viewpoint at any stage. Experiments on three benchmark cross-view human action datasets show that our method outperforms existing state-of-the-art.

\end{abstract}

}

\maketitle
\IEEEdisplaynotcompsoctitleabstractindextext

\IEEEpeerreviewmaketitle

\section{Introduction}
Video based human action recognition has many applications in human-computer interaction, surveillance, video indexing and retrieval. Actions or movements generate varying patterns of spatio-temporal appearances in videos that can be used as feature descriptors for action recognition. Based on this observation, several visual representations have been proposed for discriminative human action recognition such as space-time pattern templates~\cite{STPT2}, shape matching~\cite{SHM1,15,SHM3}, spatio-temporal interest points~\cite{Dollar,STIP,Hessian,MyECCV,STIP10,STIP11}, and motion trajectories based representation~\cite{Traj,DTraj2,DTraj1,MOTra}. Especially, dense trajectory based methods~\cite{DTraj2,DTraj1,MOTra} have shown impressive results for action recognition by tracking densely sampled points through optical flow fields. While these methods are effective for action recognition from a common viewpoint, their performance degrades significantly under viewpoint changes. This is because the same action appears different and results in different trajectories when observed from different viewpoints.

\begin{figure}[t]
\begin{center}
\includegraphics[width=8.3 cm]{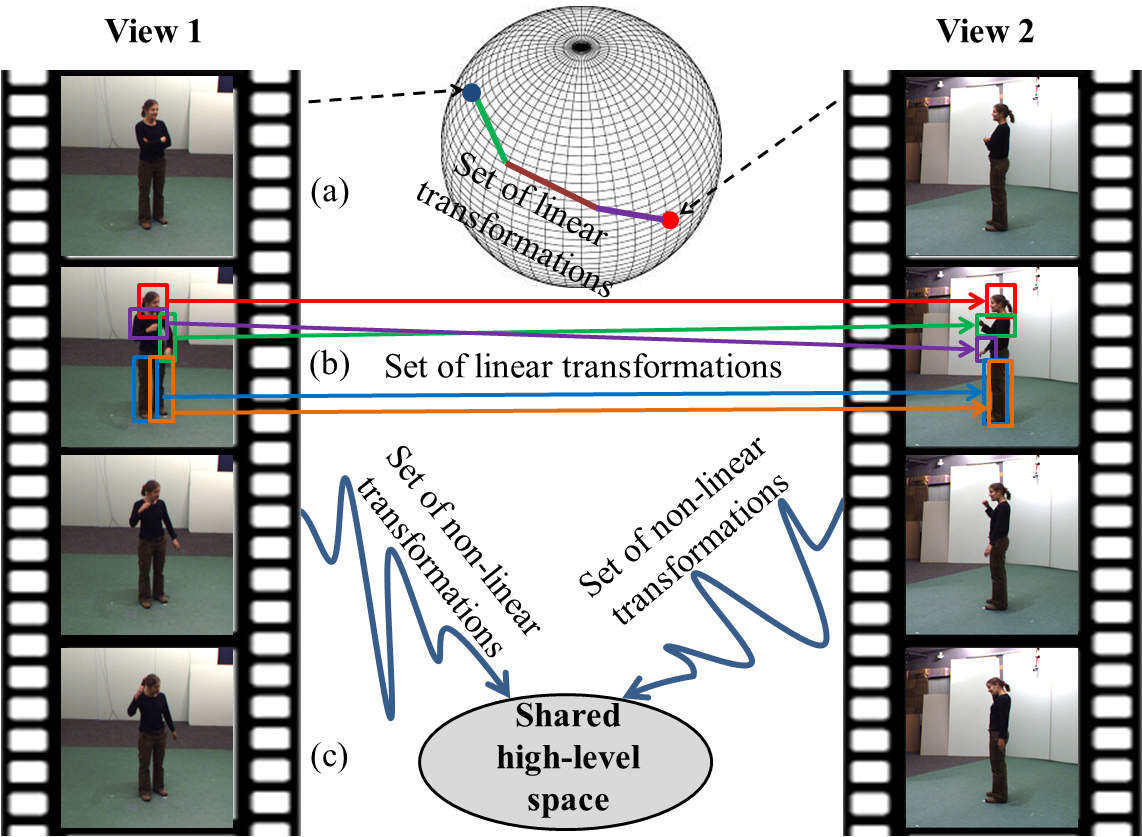}
\vspace{-3mm}
\end{center}
\caption{\small Existing cross-view action recognition techniques~\cite{CTE,CTE2,NDVV,8,7,14,and-or,CVV,virtualviews} connect two different views with a set of linear transformations that are unable to capture the non-linear manifolds on which real actions lie. (a) Li and Zickler~\cite{virtualviews} construct cross-view action descriptors by applying a set of linear transformations on view-dependent descriptors. The transformations are obtained by uniformly sampling a few points along the path connecting source and target views. (b) Wang et al.~\cite{and-or} learn a separate linear transformation for each body part using samples from training views to interpolate unseen views. (c) Our proposed R-NKTM learns a shared high-level space among all possible views. The view-dependent action descriptors from both source and target views are independently transferred to the shared space using a sequence of non-linear transformations.}
\label{fig:Motivation}
\vspace{-6mm}
\end{figure}

A practical system must recognize human actions from unknown and more importantly unseen viewpoints. One approach for recognizing actions across different viewpoints is to collect data from all possible views and train a separate classifier for each case. This approach does not scale well as it requires a large number of labelled samples for each view. To overcome this problem, some techniques infer 3D scene structure and use geometric transformations to achieve view invariance \cite{29,26,15,10,4}. These methods often require robust joint estimation which is still an open problem in real-world settings. Other methods focus on view-invariant spatio-temporal features~\cite{hankelet,17,21,IXMAS,self-sim}. However, the discriminative power of these methods is limited by their inherent structure of view-invariant features~\cite{feature}.

\begin{figure*}[t]
\centering
\includegraphics[width=17.5 cm]{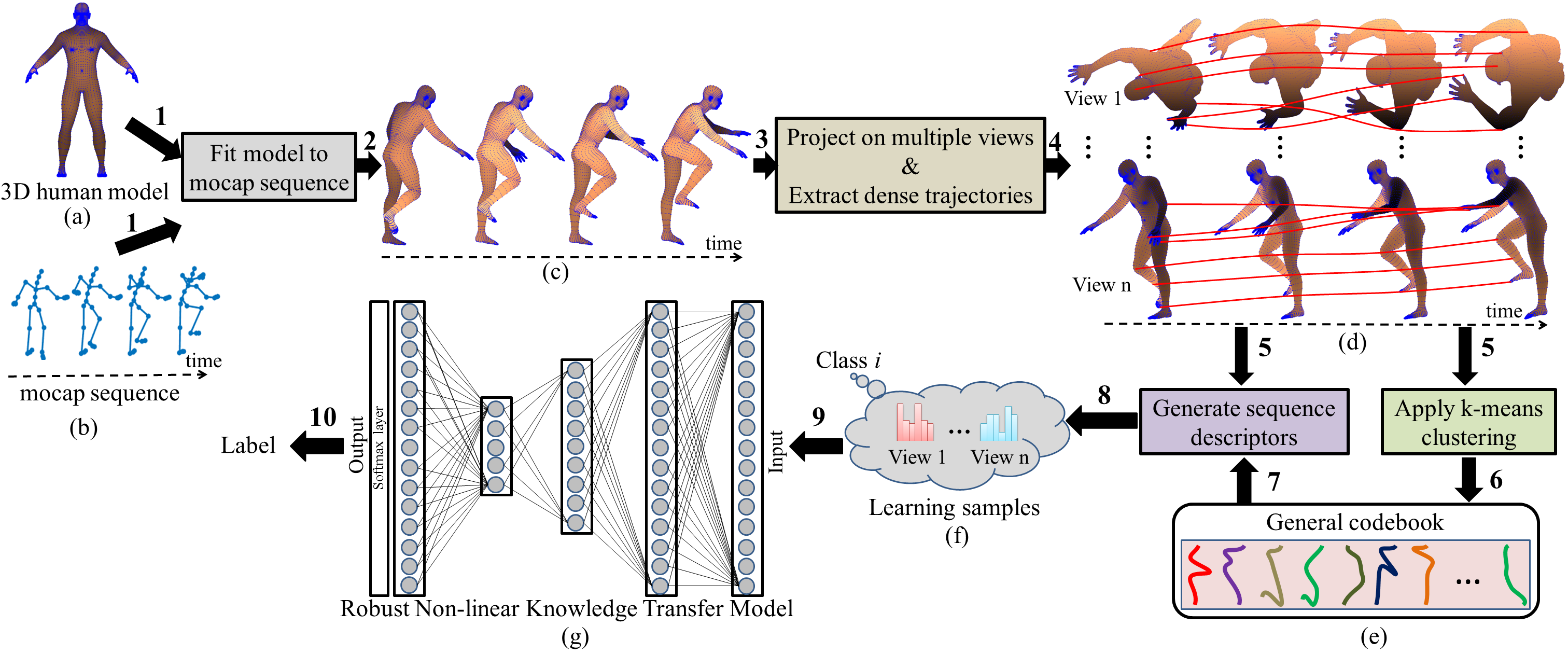}
\caption{\small Framework of the proposed R-NKTM learning algorithm. A realistic 3D human model (a) is fitted to a real mocap sequence (b) to generate 3D action video (c) which is projected to plains viewed from $n=108$ angles. Projection from only two viewpoints are shown in (d). This results in $n$ sequences of 2D pointclouds that are connected sequentially to construct synthetic trajectories (red curves in (d)) which are used to learn a general codebook (e). A bag-of-features approach is used to build the dense trajectory descriptors (f) from which a single R-NKTM (g) is learned. Note that instead of action labels, we use dummy labels where each 3D video gets a different label. The R-NKTM is learned once only and generalizes to real videos for cross-view feature extraction.}
\label{fig:Overall}
\vspace{-3mm}
\end{figure*}

Knowledge transfer-based methods~\cite{and-or,NDVV,SDVI,CVV,8,7,14,virtualviews} have recently become popular for cross-view action recognition. These methods find a view independent latent space in which features extracted from different views are directly comparable. For instance, Li and Zickler~\cite{virtualviews} proposed to construct virtual views between action descriptors from source and target views. They assume that an action descriptor transforms continuously between two viewpoints and the virtual path connecting two views lies on a hyper-sphere (see Fig.~\ref{fig:Motivation}-(a)). Thus, ~\cite{virtualviews} computes virtual views as a sequence of linearly transformed descriptors obtained by making a finite number of stops along the virtual path. This method requires samples from both source and target views during training to construct virtual views.

To relax the above constraint on training data, Wang et al.~\cite{and-or} used a set of discrete views during training to interpolate arbitrary unseen views at test time. They learned a separate linear transformation between different views for each human body part using a linear SVM solver as shown in Fig.~\ref{fig:Motivation}-(b), thereby limiting the scalability and increasing the complexity of their approach.

Existing view knowledge transfer approaches are unable to capture the non-linear manifolds where realistic action videos generally lie, especially when actions are captured from different views. This is because they only seek a set of linear transformations to construct virtual views between the descriptors of action videos captured from different viewpoints. Furthermore, such methods are either not applicable or perform poorly when recognition is performed on videos acquired from unknown and, more importantly, unseen viewpoints.

In this paper, we propose a different approach to view knowledge transfer that relaxes the assumptions on the virtual path and the requirements on the training data. We approach cross-view action recognition as a non-linear knowledge transfer learning problem where knowledge from multiple views is transferred to a shared compact high-level space. Our approach consists of three phases. Figure \ref{fig:Overall} shows an overview of the first phase where a Robust Non-linear Knowledge Transfer Model (R-NKTM) is learned. The proposed R-NKTM is a deep fully-connected network with weight decay and sparsity constraints which learns to transfer action video descriptors captured from different viewpoints to a shared high-level representation. The strongest point of our technique is that we learn a {\em single} R-NKTM for mapping all action descriptors from all camera viewpoints to a shared compact space. Note that the labels used in Fig.~\ref{fig:Overall} are dummy labels where every sequence is given a unique label that does not correspond to any specific action. Thus, action labels are not required while  R-NKTM learning or while transferring training and test action descriptors to the shared high-level space using the R-NKTM. The second phase is training where action descriptors from unknown views are passed through the learned R-NKTM to construct their cross-view action descriptors. Action labels of training data are now required to train the subsequent classifier. In the test phase, view-invariant descriptors of actions observed from unknown and previously unseen views are constructed by forward propagating their view dependent action descriptors through the learned R-NKTM. Any classifier can be trained on the cross-view action descriptors for classification in a view-invariant way. We used a simple linear SVM classifier to show the strength of the proposed R-NKTM.

Our R-NKTM learning scheme is based on the observation that similar actions, when observed from different viewpoints, still have a common structure that puts them apart from other actions. Thus, it should be possible to separate action related features from viewpoint related features. The main challenge is that these features cannot be linearly separated. The second challenge comes from learning a non-linear model itself which requires a large amount of training data. Our solution is to learn the R-NKTM from action trajectories of synthetic 3D human models fitted to real motion capture (mocap) data. By projecting these 3D human models to different views, we can generate a large corpus of synthetic trajectories to learn the R-NKTM. We use k-means to generate a general codebook for encoding the action trajectories. The same codebook is used to encode dense trajectories extracted from real action videos in the training and test phases. 

The major contribution of our approach is that we learn a {\em single} Robust Non-linear Knowledge Transfer Model (R-NKTM) which can bring any action observed from an unknown viewpoint to its compact high-level representation. Moreover, our method encodes action trajectories using a general codebook learned from synthetic data and then uses the same codebook to encode action trajectories of real videos. Thus, new action classes from real videos can easily be added using the same learned NTKM and codebook. Comparison with eight existing cross-view action recognition methods on four benchmark datasets including the IXMAS~\cite{IXMAS}, UWA3D Multiview Activity II~\cite{MyPAMI}, Northwestern-UCLA Multiview Action3D~\cite{and-or}, and UCF Sports~\cite{UCFSports} datasets shows that our method is faster and achieves higher accuracy especially when there are large viewpoint variations.

This paper is an extension of our prior work \cite{NKTM} where we transferred a given action acquired from any viewpoint to its canonical view. Knowledge of the canonical view was required for NKTM learning in \cite{NKTM}. This is a problem because the canonical view is not only action dependent, it is ill-defined. For example, what would be the canonical view of a person walking in a circle? Another limitation of \cite{NKTM} is that cylinders were fitted to the mocap data to approximate human limbs, head and torso. The trajectories generated from such models do not accurately represent human actions. In this paper, we extend our work by removing both limitations. Firstly, we no longer require identification of the canonical view for learning the new R-NKTM and use dummy labels instead. Secondly, we fit realistic 3D human models to the mocap data and hence generate more accurate trajectories. Using 3D human models also enables us to vary, and hence model, the human body shape and size. Besides these extensions, we also perform additional experiments on two more datasets namely, the UWA3D Multiview Activity II~\cite{MyPAMI} and UCF Sports~\cite{UCFSports} datasets. We denote our prior model \cite{NKTM} by NKTM and the one proposed in this paper by R-NKTM.

\vspace{-2mm}
\section{Related work}
The majority of existing literature~\cite{STPT2,SHM1,15,SHM3,Dollar,STIP,Hessian,MyECCV,STIP10,STIP11,Traj,DTraj2,DTraj1,MOTra,MyWACV14,MyPRL,ShamirPAMI,myLLC} deals with action recognition from a common viewpoint. While these approaches are quite successful in recognizing actions captured from similar viewpoints, their performance drops sharply as the viewpoint changes due to the inherent view dependence of the features used by these methods. To tackle this problem, geometry based methods have been proposed for cross-view action recognition. Rao et al.~\cite{21} introduced an action representation to capture the dramatic changes of actions using view-invariant spatio-temporal curvature of 2D trajectories. This method uses a single point (e.g.~hand centroid) trajectory. Yilmaz and Shah~\cite{29} extended this approach by tracking the 2D points on human contours. Given the human contours for each frame of a video, they generate an action volume by computing point correspondences between consecutive contours. Maximum and minimum curvatures on the spatio-temporal action volume are used as view-invariant action descriptors. However, these methods require robust interest points detection and tracking, which are still challenging problems.

Instead of using geometry constraints, Junejo et al.~\cite{self-sim} proposed Self-Similarity Matrix that is constructed by computing the pairwise similarity between any pair of frames. Hankelet~\cite{hankelet} represents actions with the dynamics of short tracklets, and achieves cross-view action recognition by finding the Hankelets that are invariant to viewpoint changes. These methods perform poorly on videos acquired from viewpoints that are significantly different from those of the training videos (e.g.~the top view of IXMAS dataset)~\cite{NKTM,14}.

Recently, transfer learning approaches have been employed to address cross-view action recognition by exploring some form of statistical connections between view-dependent features extracted from different viewpoints. A notable example of this category is the work of Farhadi et al.~\cite{8}, who employed Maximum Margin Clustering to generate split-based features in the source view, then trained a classifier to predict split-based features in the target view. Liu et. al.~\cite{14} learned a cross-view bag of bilingual words using the simultaneous multiview observations of the same action. They represented the action videos by bilingual words in both views. Zheng~\cite{SDVI} proposed to build a transferable dictionary pair by forcing the videos of the same action to have the same sparse coefficients across different views. However, these methods require feature-to-feature correspondence at the frame-level or video-level during training, thereby limiting their applications. 

Li and Zickler~\cite{virtualviews} assume that there is a smooth virtual path connecting the source and target views. They uniformly sampled a finite number of points along this virtual path and considered each point as a virtual view i.e.~a linear transformation function. Action descriptors from both views are augmented into cross-view feature vectors by applying a finite sequence of linear transformations to each descriptor. Recently, Zhang et al.~\cite{CVV} extended this approach by applying an infinite sequence of linear transformations. Although these methods can operate in the absence of feature-to-feature correspondence between source and target views, they still require the samples from target view during training.

More recently, Wang et al.~\cite{and-or} proposed cross-view action recognition by discovering discriminative 3D Poselets and learning the geometric relations among different views. However, they learn a separate transformation between different views using a linear SVM solver. Thus many linear transformations are learned for mapping between different views. For action recognition from unseen views, all learned transformations are used for exhaustive matching and the results are combined with an AND-OR Graph (AOG). This method also requires 3D skeleton data for training which is not always available. Gupta et al.~\cite{CTE} proposed to find the best match for each training video in large mocap sequences using a Non-linear Circular Temporary Encoding method. The best matched mocap sequence and its projections on different angles are then used to generate more synthetic training data making the process computationally expensive. Moreover, the success of this approach depends on the availability of a large mocap dataset which covers a wide range of human actions~\cite{CTE, CTE2}. \\

\vspace{-3mm} \noindent {\bf Deep Learning Models:} Deep learning models~\cite{DLM1,DLM2,DLM3} can learn a hierarchy of features by constructing high-level representations from low-level ones. Due to the impressive results of such deep learning on handwritten digit recognition~\cite{DLM2}, image classification~\cite{CNNimage} and object detection~\cite{CNNobject}, several methods have been recently proposed to learn deep models for video based action recognition. Ji et al.~\cite{C3D} extended the deep 2D convolutional neural network (CNN) to 3D where convolutions are performed on 3D feature maps from spatial and temporal dimensions. Simonyan and Zisserman~\cite{2CNN} trained two CNNs, one for RGB images and one for optical flow signals, to learn spatio-temporal features. Gkioxari and Malik~\cite{AT} extended this approach for action localization. Donahue et al.~\cite{LRCN} proposed an end-to-end trainable recurrent convolutional network which processes video frames with a CNN, whose outputs are passed through a recurrent neural network. None of these methods is designed for action recognition in videos acquired from unseen views. Moreover, learning deep models for the task of cross-view action recognition requires a large corpus of training data acquired from multiple views which is unavailable and very expensive to acquire and label. These limitations motivate us to propose a pipeline for generating realistic synthetic training data and subsequently learn a Robust Non-linear Knowledge Transfer Model (R-NKTM) which can transfer action videos from any view to a high level space where actions can be matched in a view-invariant way. Although learned from synthetic data, the proposed R-NKTM is able to generalize to real action videos and achieve state-of-the-art results.

\section{Proposed technique}
\label{ProposedTechnique}
The proposed technique comprises three main stages including feature extraction, Robust Non-linear Knowledge Transfer Model (R-NKTM) learning, and cross-view action description. In the feature extraction stage, synthetic dense trajectories are first generated by fitting 3D human models to mocap sequences and projecting the resulting 3D videos on plains corresponding to different viewpoints. The 2D dense trajectories are then represented by bag-of-features. In the model learning stage, a deep fully-connected network, called R-NKTM, is learned such that it transfers the view-dependent trajectory descriptors of the same action observed from different viewpoints to a shared high-level virtual view. In the third stage, the dense trajectory descriptors of real action videos are passed through the learned R-NKTM to construct cross-view action descriptors. Details of each stage are given below.

\vspace{-3mm}
\subsection{Feature extraction}
\label{Sec:FeatureExtraction}
Dense trajectories have shown to be effective for action recognition~\cite{MOTra,DTraj2,DTraj1,CTE}. Our motivation for using dense trajectories is that they can be easily extracted from conventional videos as well as the synthetic 3D videos generated from mocap data.
 
\vspace{-2mm}
\subsubsection{Dense trajectories from videos}
To extract trajectories from videos, Wang et al.~\cite{DTraj1,DTraj2} proposed to sample dense points from each frame and track them using displacement information from a dense optical flow field.  The shape of a trajectory encodes the local motion pattern. Given a trajectory of length $L$, a sequence $S$ of displacement vectors $\Delta P_t=(P_{t+1}-P_t)=(x_{t+1}-x_t,y_{t+1}-y_t)$ is formed and normalized as:
\begin{equation}
S =\frac{(\Delta P_t,...,\Delta P_{t+L-1})}{\sum_{i=t}^{t+L-1}{\|\Delta P_i\|}}.
\label{NS}\end{equation}
The descriptor $S$ encodes the shape of the trajectory. To embed appearance and motion information, a spatio-temporal volume aligned with the trajectory is subdivided into a spatio-temporal grid and HOG, HOF and MBH descriptors are computed in each cell of the grid. The bag-of-features approach is then employed to construct a histogram of visual word occurrences for each descriptor (trajectory shape, HOG, HOF, MBH) separately. The final descriptor is a concatenation of these four histograms. However, it is important to note that unlike~\cite{DTraj1,DTraj2} we only use the trajectory descriptors since their extraction using multiple viewpoints and scales is computationally efficient as shown in Section~\ref{sec:DTmocap}. The same process, on the other hand, is computationally very expensive for the remaining three descriptors i.e. HOG, HOF, and MBH. Moreover, using trajectories only is also robust to changes in visual appearance due to clothing and lighting conditions.

\vspace{-2mm}
\subsubsection{Dense trajectories from mocap sequences}
\label{sec:DTmocap}
Figure \ref{fig:Overall} gives an overview of the steps involved in generating synthetic dense trajectories using  different human body shapes performing a large number of actions rendered from numerous viewpoints. Details are below.\\

\vspace{-3mm}\noindent {\bf 3D Human body models:} There are different ways to generate 3D human models. For example, Bogo et al.~\cite{FAUST} developed the FAUST dataset containing full 3D human body scans of $10$ individuals in $30$ poses. However, the skeleton data is not provided for these scans. Another way to generate a 3D human model is to use the open source MakeHuman software~\cite{MakeHuman} which can synthesize different realistic 3D human shapes in a predefined pose and also provide the joints positions which can be used for generating human models in different poses. We use this technique for generating the 3D human models in our work.\\ 

\vspace{-3mm}\noindent {\bf Fitting 3D human models to mocap sequences:} Several approaches~\cite{MOSH,SCAPE} have been proposed in the literatures to fit a 3D human model to the motion capture skeleton data of a human subject. For instance, the SCAPE method~\cite{SCAPE} learns pose and body-shape deformation models from the training scans of different human bodies in a few poses. Given a set of markers, SCAPE constructs a full mesh which is consistent with the SCAPE models, best matches with the given markers and maintains realistic muscle deformations. This method takes approximately $3$ minutes to generate each frame. Another example is the MoSh method~\cite{MOSH} which estimates an accurate 3D body shape directly from the mocap skeleton without the use of 3D human scans. MoSh is also able to estimate soft-tissue motions from mocap data and subsequently use them to produce animations with subtlety and realism. MoSh requires about $7$ minutes to estimate a subject's shape. However, these methods are computationally expensive to apply on a large corpus of mocap sequences. Thus, we use the open source Blender package~\cite{Blender} to fit 3D human models to mocap data. Given a 3D human model generated by the MakeHuman software and a mocap sequence, Blender normalizes the mocap skeleton data with respect to the skeleton data of the human model and then fits the model to the normalized mocap data. This process results in a synthetic but realistic full 3D human body video corresponding to a mocap sequence. \\

\begin{figure}
\begin{center}
\includegraphics[width=4cm]{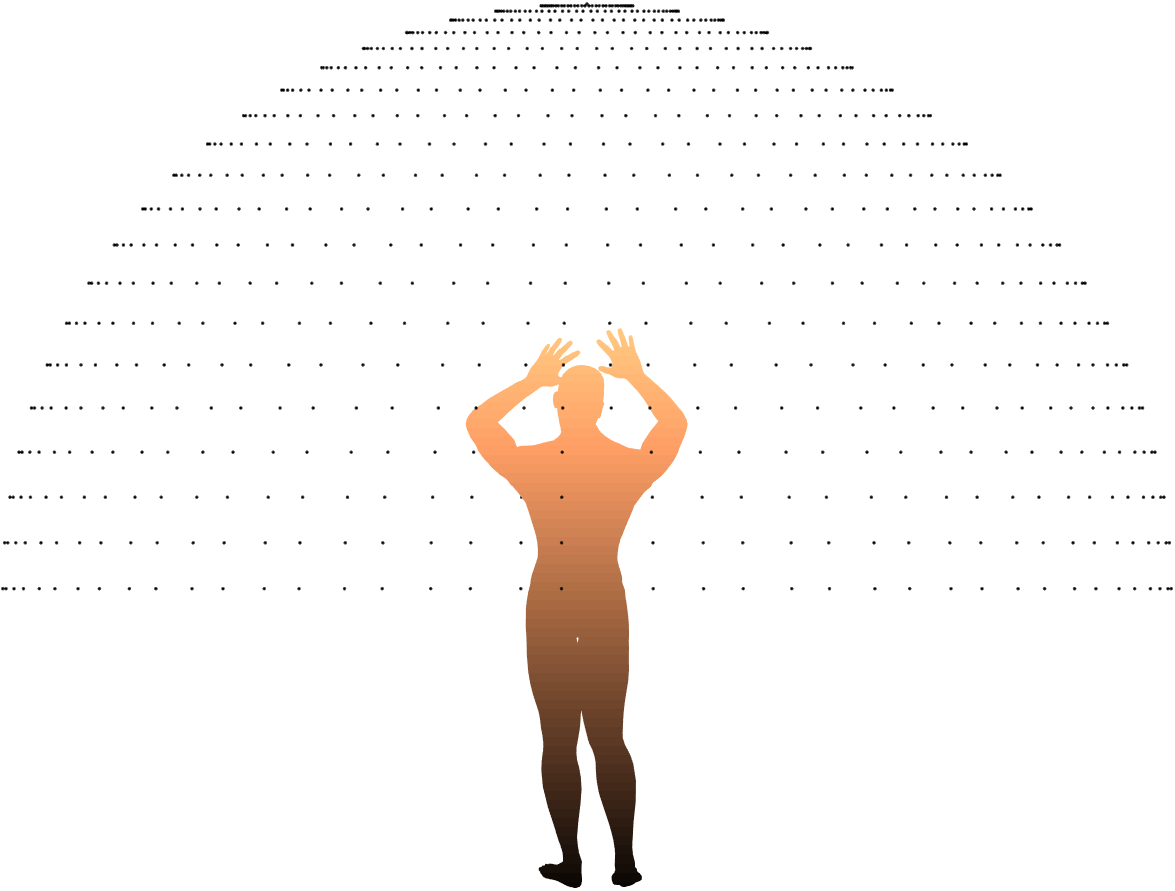}
\end{center}
\caption{\small Virtual cameras are placed on the hemisphere looking towards the center of the sphere to generate 108 virtual views.}
\vspace{-5mm}
\label{fig:virtualcameras}
\end{figure}

\vspace{-3mm}\noindent {\bf Projection from multiple viewpoints:} We deploy a total of $108$ synthetic cameras (at distinct latitudes and longitudes) on a sphere surrounding the subject performing an action, as shown in Fig.~\ref{fig:virtualcameras}. Given a perspective camera and a frame of a synthetic full 3D human body sequence, we deal with self-occlusions by removing points that are not visible from the given camera viewpoint. First, we perform back-face culling by removing 3D points which their normals face away from the camera. Then, the hidden point removal technique~\cite{HPR} is applied on the remaining 3D points. This gives us a set of visible 3D points corresponding to the given viewpoint. The visible 3D points are projected to the $x-y$ plain using perspective projection resulting in a 2D pointcloud. We repeat this process for all $108$ cameras and all frames of the synthetic full 3D human body sequence, thereby, $108$ sequences of 2D pointclouds are generated for each synthetic full 3D human action sequence corresponding to a mocap sequence.\\

\vspace{-3mm}\noindent {\bf Dense trajectory extraction:} Since we already have dense correspondence between the 3D human models in each pose, it is straight forward to extract trajectory features from their projected sequence of 2D pointclouds by simply connecting them in time over a fixed horizon of $L$ frames. A sequence $S$ of normalized displacement vectors $\Delta P_t$ is calculated for each point~\eqref{NS}. Note that we use the same $L=15$ for both synthetic and real videos. We represent each video (synthetic or real) by a set of motion trajectory descriptors. We construct a codebook of size $k=2000$ by clustering the trajectory descriptors with $k$-means. It is important to note that clustering is performed only over the synthetic trajectory descriptors to learn the codebook. Thus, unlike existing cross-view action recognition techniques~\cite{CTE,CTE2,virtualviews,NDVV,14} the codebook we learn does not use the trajectory descriptors of real videos from IXMAS~\cite{IXMAS}, UWA3DII~\cite{MyPAMI} or Northwestern-UCLA~\cite{and-or} datasets. We call this the general codebook. We consider each cluster as a codeword that represents a specific motion pattern shared by the trajectory descriptors in that cluster. One codeword is assigned to each trajectory descriptor based on the minimum Euclidean distance. The resulting histograms of codeword occurrences are used as trajectory descriptors. Real action videos are encoded with the same codebook. Recall that unlike dense trajectory-based methods~\cite{DTraj2,DTraj1} which use HOF, HOG, and MBH descriptors along with trajectories, our method only uses trajectory descriptors.

\vspace{-3mm}
\subsection{Non-linear Knowledge Transfer Model}
Besides the limitations of employing linear transformation functions between views, existing cross-view action recognition methods~\cite{and-or,CTE,8,14,CVV,virtualviews} are either not applicable to unseen views or require augmented training samples which cover a wide range of human actions. Moreover, these methods do not scale well to new data and need to repeat the computationally expensive model learning process when a new action class is to be added. To simultaneously overcome these problems, we propose a Robust Non-linear Knowledge Transfer Model (R-NKTM) that learns to transfer the action trajectory descriptors from all possible views to a shared compact high-level virtual view. Our R-NKTM is learned using synthetic training data and is able to generalize to real data without the need for retraining or fine-tuning, thereby increasing its scalability.

\begin{figure}
\centering
\includegraphics[width=8.3 cm]{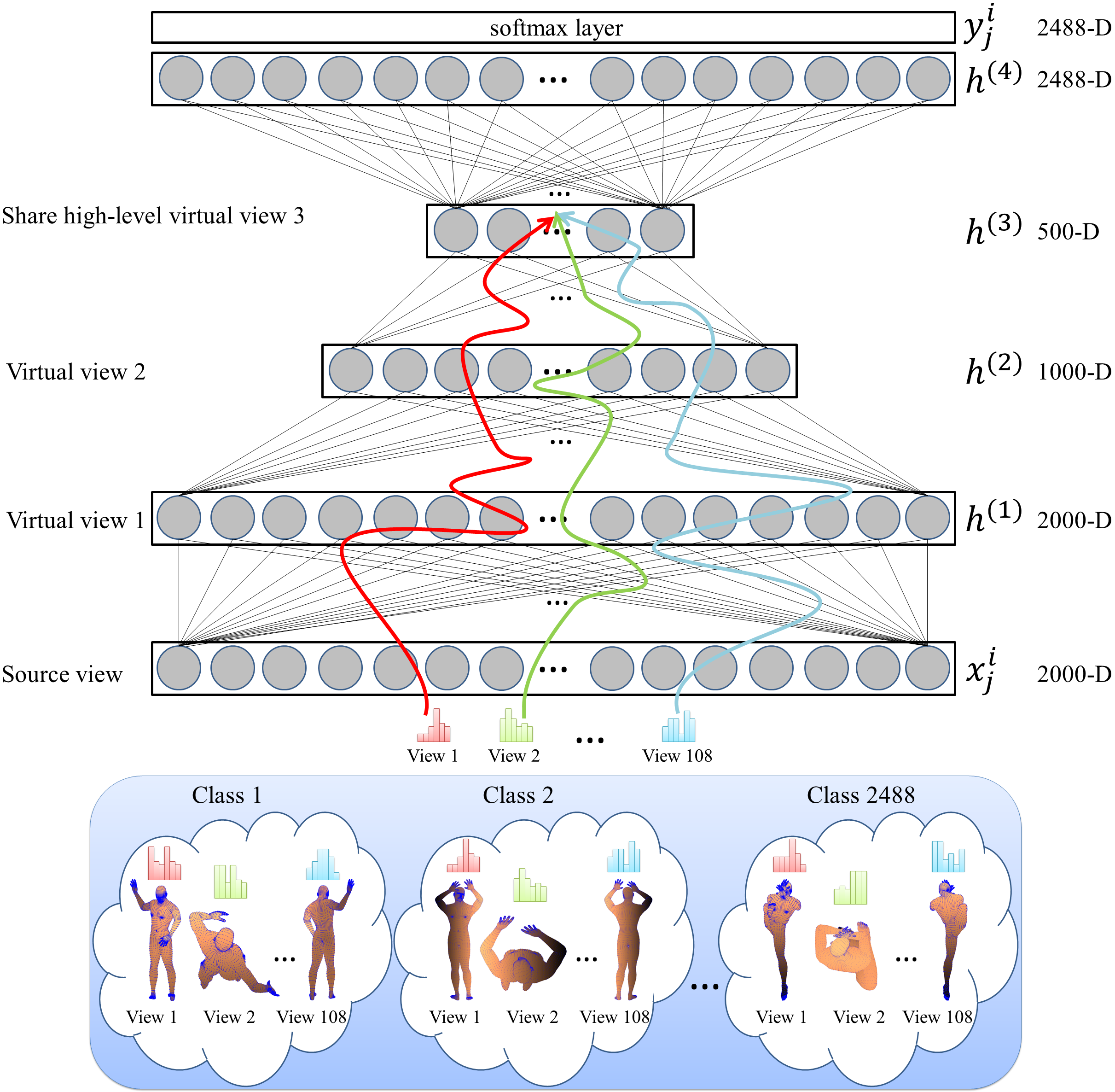}
\caption{\small Assume that there are $n$ virtual paths connecting $n$ input views to a shared high-level virtual view. We show only $3$ different virtual paths for $3$ views. Our R-NKTM learns to find this shared high-level space and the non-linear virtual paths connecting input views to this shared space. Dummy labels are used to learn the model i.e.~every video is given a different unique label.}
\vspace{-6mm}
\label{fig:TrainingPhase}
\end{figure}

As depicted in Fig.~\ref{fig:TrainingPhase}, our R-NKTM is a deep network, consisting of $Q$ fully-connected layers (where $Q=4$) followed by a softmax layer and $p^{(q)}$ units in the $q$-th fully-connected layer where $q=1,2,\cdots,Q$ and $p^{(1)}=2000, p^{(2)}=1000, p^{(3)}=500, p^{(4)}=2488$. For a given training sample ${\bf x}_j^i \in \mathds{R}^{k}$, where ${\bf x}_j^i$ is the $j$-th sample in $i$-th view, the output of the first layer is ${\bf h}^{(1)}=f({\bf W}^{(1)}x_j^i+{\bf b}^{(1)}) \in \mathds{R}^{p^{(1)}}$, where ${\bf W}^{(1)} \in \mathds{R}^{p^{(1)}\times k}$ is a weight matrix to be learned in the first layer, ${\bf b}^{(1)} \in \mathds{R}^{p^{(1)}}$ is a bias vector, and $f(\cdot)$ is a non-linear activation function which is typically a ReLU (Rectified Linear Unit), sigmoid or tangent hyperbolic function. The ReLU function, $f(a)=\max(0,a)$, does not suffer from the gradient vanishing problem like the sigmoid and tangent hyperbolic functions do. Moreover, it has been shown that deep networks can be trained efficiently using the ReLU function even without the need for pre-training~\cite{ReLU}. Finally, ReLU generates sparse representations with true zeros that are suitable for exploiting sparsity in the data which is the case for histogram of codeword occurrences~\cite{ReLU}. Therefore, we use ReLU as the activation function in our proposed model. 

The output of the first layer ${\bf h}^{(1)}$ is used as the input of the second layer. The output of the second layer is computed as :
\begin{equation}
{\bf h}^{(2)}=f({\bf W}^{(2)}{\bf h}^{(1)}+{\bf b}^{(2)}) \in \mathds{R}^{p^{(2)}}, 
\end{equation}
where ${\bf W}^{(2)} \in \mathds{R}^{p^{(2)} \times p^{(1)}}$, ${\bf b}^{(2)} \in \mathds{R}^{p^{(2)}}$, and $f(\cdot)$ are the weight matrix, bias, and non-linear activation function of the second layer, respectively. Similarly, the output of the second layer ${\bf h}^{(2)}$ is used as the input of the third layer and the output of the third layer is computed as
\begin{equation}
{\bf h}^{(3)}=f({\bf W}^{(3)}{\bf h}^{(2)}+{\bf b}^{(3)}) \in \mathds{R}^{p^{(3)}}, 
\end{equation}
where ${\bf W}^{(3)} \in \mathds{R}^{p^{(3)} \times p^{(2)}}$, ${\bf b}^{(3)} \in \mathds{R}^{p^{(3)}}$, and $f(\cdot)$ are the weight matrix, bias, and non-linear activation function of the third layer, respectively. The output of the last fully-connected layer is computed as
\begin{equation}
g({\bf x}_j^i)={\bf h}^{(Q)}=f({\bf W}^{(Q)}{\bf h}^{(Q-1)}+{\bf b}^{(Q)}) \in \mathds{R}^{p^{(Q)}}
\end{equation}
where $g(\cdot)$ is a non-linear transformation function determined by the parameters ${\bf W}^{(q)}$ and ${\bf b}^{(q)}$. The output of the last fully-connected layer ${\bf h}^{(Q)}$ is passed through a softmax layer to find the appropriate class label. 

\begin{figure*}[tbp!]
\centering
\includegraphics[width=\linewidth]{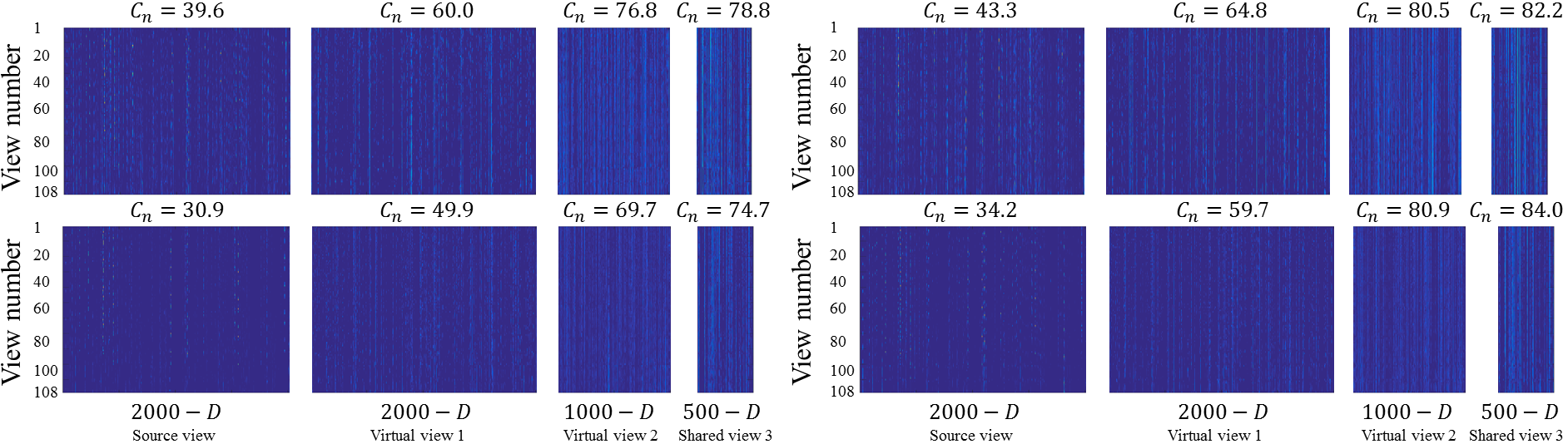}
\caption{\small Visualization of R-NKTM layer outputs for four unseen mocap sequences. Each sequence gives $108$ descriptors corresponding to different views. The outputs of the R-NKTM layers (source view ${\bf x}^i_j$ and three virtual views ${\bf h}^{(1)}$, ${\bf h}^{(2)}$, ${\bf h}^{(3)}$) are visualized as images. The 108 rows in an image correspond to 108 viewpoints of the same action. The norm of correlation coefficient ($C_n$) is shown above each image where larger values indicate higher similarity between the rows. Note that as the action descriptors progress through the R-NKTM layers, the similarity of the same action observed from 108 viewpoints increases.}
\vspace{-4mm}
\label{fig:Layers_images}
\end{figure*}

We use this structure to find a shared high-level space among all possible views. Specifically, in our problem, the inputs to the R-NKTM are synthetic trajectory descriptors corresponding to mocap sequences over different views, while the output is their dummy class labels. Since we use the CMU mocap dataset~\cite{mocap} consisting of $2488$ action sequences, the last fully-connected layer has $2488$ units whose outputs are given to the softmax layer. The basic idea of this R-NKTM is that regardless of the input view of an unknown action (recall that we do not use the action labels of the mocap sequences), we encourage the output class label of the R-NKTM to be the same for all views of the given action. We explain this idea in the following.

Assume that there is a virtual path which connects any view to a single shared high-level virtual view. Therefore, there are $n$ different virtual paths connecting $n$ input views to the shared virtual view as shown in Fig.~\ref{fig:TrainingPhase}. We consider each virtual path as a set of non-linear transformations of action descriptors. Moreover, assume that the videos of the same action over different views share the same high-level feature representation. Given these two assumptions, our objective is to find this shared high-level virtual view and the intermediate virtual views connecting the input views to the shared virtual view. 

The learning of the proposed R-NKTM is carried out by updating its parameters $\theta_{K}=\{\theta_{\bf W},\theta_{\bf b}\}$, where $\theta_{\bf W}=\{{\bf W}^{(1)},{\bf W}^{(2)},\cdots,{\bf W}^{(Q)}\}$ and $\theta_{\bf b}=\{{\bf b}^{(1)},{\bf b}^{(2)},\cdots,{\bf b}^{(Q)}\}$, for minimizing the following objective function over all samples of the input views:
\begin{equation}
E_1(\theta_{K};{\bf x}_j^i \in {\bf X})=\frac{1}{2nm}\sum_{j=1}^{m}{\sum_{i=1}^{n}{\ell(z_j,g({\bf x}_j^i))}}
\label{eq:objectiveF}\end{equation}
where $n$ is the number of viewpoints, $m$ is the number of samples in the mocap dataset (for CMU mocap dataset~\cite{mocap}: $m=2488$), $z_j$ denotes class label of the $j$-th mocap sequence i.e. $z_j=j$ and $\ell$ denotes softmax loss function. 

Due to the high flexibility of the proposed R-NKTM (e.g.~number of units in each layer $p^{(q)}$, $\theta_{K}$), appropriate settings in the configuration of the R-NKTM are needed to ensure that it learns the underlying data structure. Since the input data ${\bf x}_j^i \in \mathds{R}^{p^{(0)}}$, where $p^{(0)}=2000$, we discard the redundant information in the high dimensional input data by mapping it to a compact, high-level and low dimensional representation. This operation is performed by $3$ fully-connected layers (${\bf h}^{(1)}, {\bf h}^{(2)}, {\bf h}^{(3)}$) of the R-NKTM. 

To avoid over-fitting and improve generalization of the R-NKTM, we add weight decay $J_{w}$ and sparsity $J_{s}$ regularization terms to the training criterion i.e.~the loss function \eqref{eq:objectiveF}~\cite{ovfit1,ovfit2}. Large weights cause highly curved non-smooth mappings. Weight decay keeps the weights small and hence the mappings smooth to reduce over-fitting~\cite{smooth}. Similarly, sparsity helps in selecting the most relevant features to improve generalization.
\begin{equation}
E_2(\theta_{K};{\bf x}_j^i \in {\bf X})=E_1(\theta_{K};{\bf x}_j^i \in {\bf X})+\lambda_{w}J_{w}+\lambda_{s}J_{s}
\label{eq:objectiveFR}\end{equation}
where $\lambda_{w}$ and $\lambda_{s}$ are the weight decay and sparsity parameters respectively. The $J_{w}$ penalty tends to decrease the magnitude of the weights $\theta_{\bf W}=\{{\bf W}_1,{\bf W}_2,{\bf W}_3\}$:
\begin{equation}
J_{w}=\sum_{q=1}^Q{\|{\bf W}^{(q)}\|_F^2},
\end{equation}
where $\|{\bf W}^{(q)}\|^2_F$ returns the Frobenius norm of the weight matrix ${\bf W}^{(q)}$ of the $q$-th layer. Let 
\begin{equation}
\hat{\rho}_t^{(q)}=\frac{1}{M}\sum_{i=1}^{n}{\sum_{j=1}^{m_i}{{\bf h}_t^{(q)}({\bf x}_j^i)}}~,
\end{equation}
be the mean activation of the $t$-th unit of the $q$-th layer (averaged over all the training samples ${\bf x}_j^i \in {\bf X}$). The $J_{s}$ penalty forces the $\hat{\rho}_t^{(q)}$ to be as close as possible to a sparsity target $\rho$ and is defined in terms of the Kullback-Leibler (KL) divergence between a Bernoulli random variable with mean $\hat{\rho}_t^{(q)}$ and a Bernoulli random variable with mean $\rho$ as
\begin{equation}
\begin{split}
J_{s}&=\sum_{q=1}^{Q}{\sum_{t}{{\text{KL}}(\rho \| \hat{\rho}_t^{(q)})}} \\ 
&=\sum_{q=1}^{Q}{\sum_{t}{\rho \log{\frac{\rho}{\hat{\rho}_t^{(q)}}}}+(1-\rho)\log{\frac{1-\rho}{1-\hat{\rho}_t^{(q)}}}}~.
\end{split}
\end{equation}

The reasons for using these two regularization terms are twofold. Firstly, not all features are equally important. Secondly, sparsity forces the R-NKTM to find a compact, shared and high-level virtual view, ${\bf h}^{(3)}$, by selecting only the most critical features. A dense representation may not learn a good model because almost any change in the input layer modifies most of the entries in the output layer.

Our goal is to solve the optimization problem $E_2(\theta_{K};{\bf x}_j^i \in {\bf X})$ in \eqref{eq:objectiveFR} as a function of $\theta_{\bf W}$ and $\theta_{\bf b}$. Therefore, we use stochastic gradient descent through back-propagation to minimize this function over all training samples in the mocap data ${\bf x}^i_j \in {\bf X}$. 

Figure~\ref{fig:Layers_images} visualizes the output features of the learned R-NKTM layers for four mocap actions that were not used during learning. In each case, a 3D human model was fitted to the mocap sequence and projected  from $108$ viewpoints. Dense trajectories of each view were calculated to get $108$ descriptors which were then individually passed through the learned R-NKTM. Figure \ref{fig:Layers_images} shows the outputs of each layer as an image. As expected, the outputs of the shared virtual view ${\bf h}^{(3)}$ are very similar for all $108$ views. Note that we drop the outputs of the last fully-connected ${\bf h}^{(4)}$ and softmax layers because they are the $2048$ class scores which correspond to dummy labels.

\begin{figure*}[t!]
\centering
\includegraphics[width=17 cm]{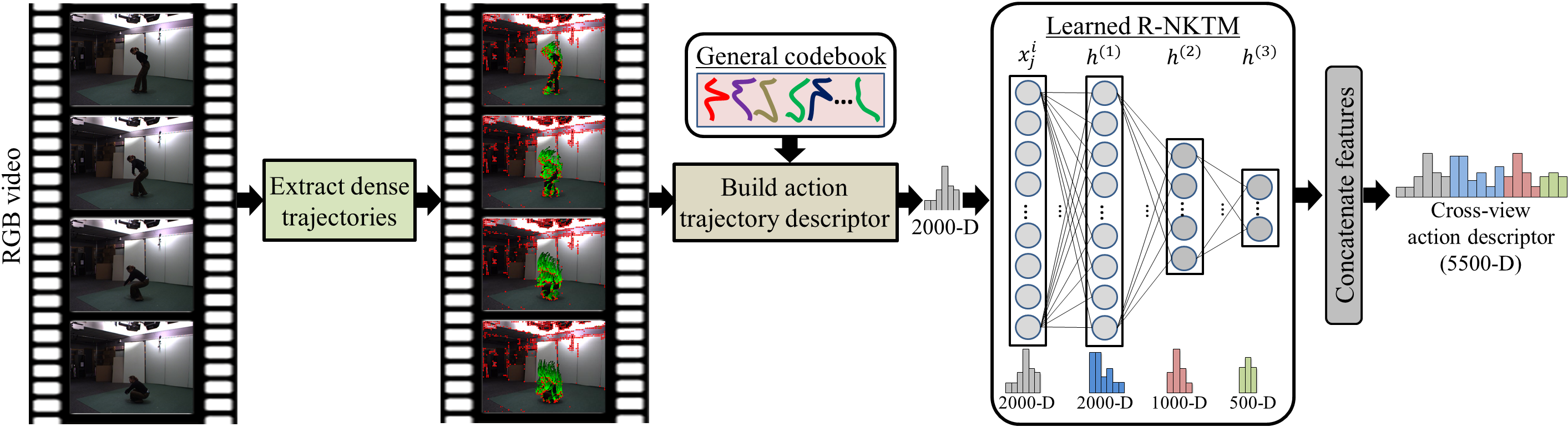}
\vspace{-2mm}
\caption{\small Extracting cross-view action descriptors from real videos. The view-dependent dense trajectory descriptor ${\bf x}$ is extracted from a training or test video and forward propagated through the learned R-NKTM for transfer to the shared high-level virtual view by performing a set of non-linear transformations. Each transformation results in a virtual view lying on the non-linear virtual path connecting the input and shared virtual views. The outputs of these transformation functions $\{{\bf v}_S,{\bf v}_1,{\bf v}_2,{\bf v}_3\}$ are concatenated to form a cross-view action descriptor. Note that the last fully-connected ${\bf h}^{(4)}$ and the softmax layers of the R-NKTM are removed during feature extraction because they correspond to the dummy labels used during R-NKTM learning.}
\vspace{-3mm}
\label{fig:TrainingPhaseSVM}
\end{figure*}

\vspace{-3mm}
\subsection{Cross-View Action Description}
\label{CVActDes}
So far we have learned an R-NKTM whose input is a synthetic trajectory descriptor corresponding to a mocap sequence fitted with a 3D human model and observed from any arbitrary view. The output of the model is the class label which is the same for all views of the sequence. However, our aim is to extract cross-view action descriptors from real videos acquired from any arbitrary view. 

Given a real human action video, the view-dependent descriptor ${\bf x}$ is constructed by extracting dense trajectories from multiple spatial scales of the given video and then building the histogram of codeword occurrences using the learned general codebook as discussed in Section~\ref{Sec:FeatureExtraction}. Recall that the R-NKTM learns to find a shared high-level virtual view, ${\bf h}^{(3)}$, and the intermediate virtual views, ${\bf h}^{(1)},{\bf h}^{(2)}$, lie on the virtual path connecting the input view and the shared virtual view. This means that we have a set of non-linear transformation functions which transfer the view-dependent action trajectory descriptor ${\bf x}$ from an unknown view to the shared high-level virtual view. Recall that we remove the last fully-connected ${\bf h}^{(4)}$ and softmax layers because these layers correspond to dummy labels which do not provide any useful information for representing real videos.


We describe an action video as alterations of its view-dependent descriptor along the virtual path. The cross-view action descriptor is constructed by concatenating the transformed features along the virtual path into a long feature vector $\left[{\bf x}, {\bf h}^{(1)}, {\bf h}^{(2)}, {\bf h}^{(3)} \right]$. This new descriptor implicitly incorporates the non-linear changes from the unknown input view to the shared high-level virtual view. Since the feature vector contains all the virtual views from the source to the shared view, it is more robust to viewpoint variations. To perform cross-view action recognition on any real action video dataset, we use the samples with their corresponding labels from a source view i.e.~training data, and extract their cross-view action descriptors. Then, we train a linear SVM classifier to classify these actions. For a given sample at test time (i.e.~samples from target view), we simply extract its cross-view descriptor and feed it to the trained SVM classifier to find its label. Figure~\ref{fig:TrainingPhaseSVM} shows an overview of the proposed method for extracting cross-view action descriptors from real videos.

\section{Experiments}
We evaluate our proposed method on four benchmark datasets including the INRIA Xmas Motion Acquisition Sequences (IXMAS)~\cite{IXMAS}, UWA3D Multiview Activity3DII (UWA3DII)~\cite{MyPAMI}, Northwestern-UCLA Multiview Action3D (N-UCLA)~\cite{and-or}, and UCF Sports~\cite{UCFSports} datasets. We compare our performance to the state-of-the-art action recognition methods including Dense Trajectories (DT)~\cite{DTraj1}, Hankelets~\cite{hankelet}, Discriminative Virtual Views (DVV)~\cite{virtualviews}, Continuous Virtual Path (CVP)~\cite{CVV}, Non-linear Circulant Temporal Encoding (nCTE)~\cite{CTE}, AND-OR Graph (AOG)~\cite{and-or}, Long-term Recurrent Convolutional Network (LRCN)~\cite{LRCN}, and Action Tube\cite{AT}.  

We report action recognition results of our method for unseen and unknown views i.e.~unlike DVV~\cite{virtualviews} and CVP~\cite{CVV} we assume that no videos, labels or correspondences from the target view are available at training time. More importantly, unlike existing techniques~\cite{virtualviews,CVV,CTE,and-or,AT,LRCN} we learn our R-NKTM and build the codebook using only synthetic motion trajectories generated from mocap sequences. Therefore, the R-NKTM and the codebook are general and can be used for cross-view action recognition on any action video without the need for retraining or fine-tuning. More precisely, we use the same learned R-NKTM to evaluate our algorithm on IXMAS~\cite{IXMAS}, UWA3DII~\cite{MyPAMI} and N-UCLA~\cite{and-or} datasets. However, nCTE~\cite{CTE}, DVV~\cite{virtualviews}, CVP~\cite{CVV} and AOG~\cite{and-or} need to learn different models to transfer knowledge across views for different datasets. Action Tube~\cite{AT} and LRCN~\cite{LRCN} require to fine-tune a pre-trained model for each action video dataset.

\begin{figure}[t]
\centering
\includegraphics[width=8.3 cm]{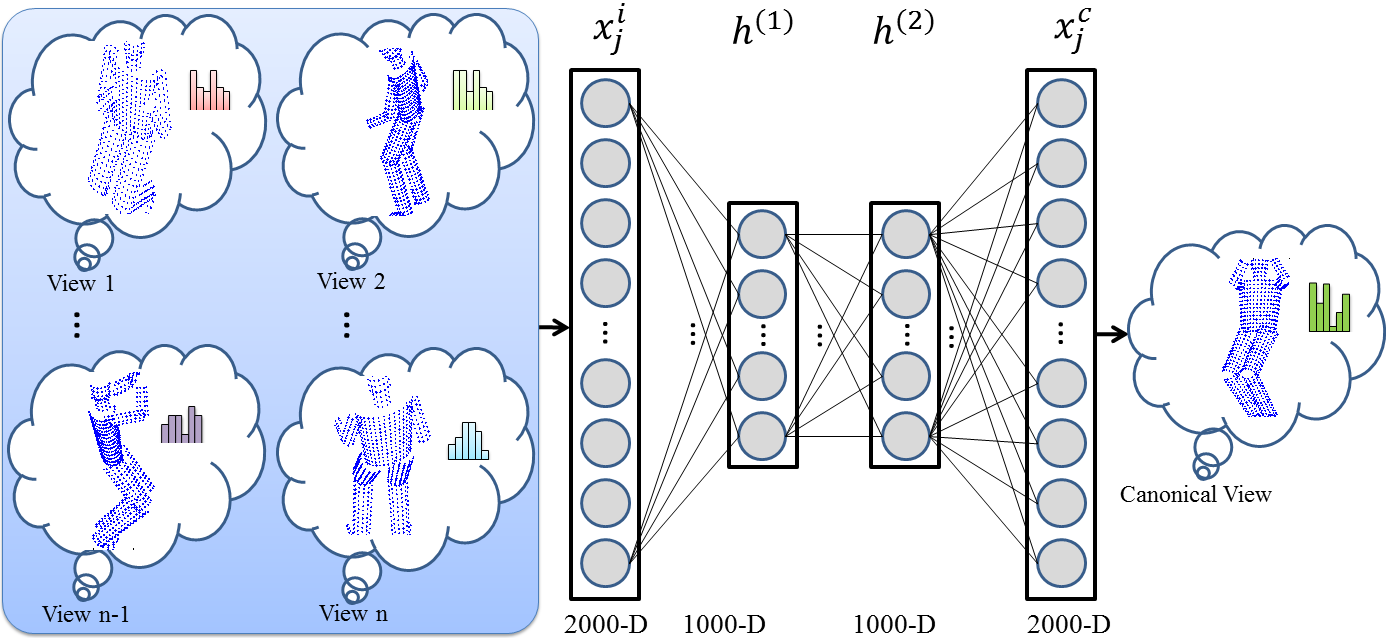}
\vspace{-1mm}
\caption{\small NKTM~\cite{NKTM} learns to bring any action observed from an unknown viewpoint to its canonical view. NTKM architecture is different from the R-NKTM proposed in this paper (see Fig.~\ref{fig:TrainingPhase}).}
\vspace{-5mm}
\label{fig:NKTM1}
\end{figure}

In addition to the accuracy of our method, we report the recognition accuracy of the NKTM proposed in our prior work~\cite{NKTM}. As shown in Fig.~\ref{fig:NKTM1}, the view knowledge transfer model in~\cite{NKTM} uses a different architecture consisting of $2000$ units at the input/output layers and $1000$ units at the two hidden layers. Moreover, it learns to transfer actions observed from unknown viewpoints to their canonical view. 

\vspace{-3mm}
\subsection{Implementation Details}
\label{details}
For a fair comparison, we pass the dense trajectory descriptors, instead of spatio-temporal interest point descriptors, to DVV~\cite{virtualviews} and CVP~\cite{CVV}. Moreover, we use $10$ virtual views, each with a $30$-dimensional features. The baseline results are obtained using publicly available implementations of DT~\cite{DTraj1}, Hankelets~\cite{hankelet}, nCTE~\cite{CTE}, DVV~\cite{virtualviews}, LRCN~\cite{LRCN} and Action Tube~\cite{AT} or from the original papers. \\

\begin{figure}[t]
\centering
\vspace{3mm}
\includegraphics[width=8.3 cm]{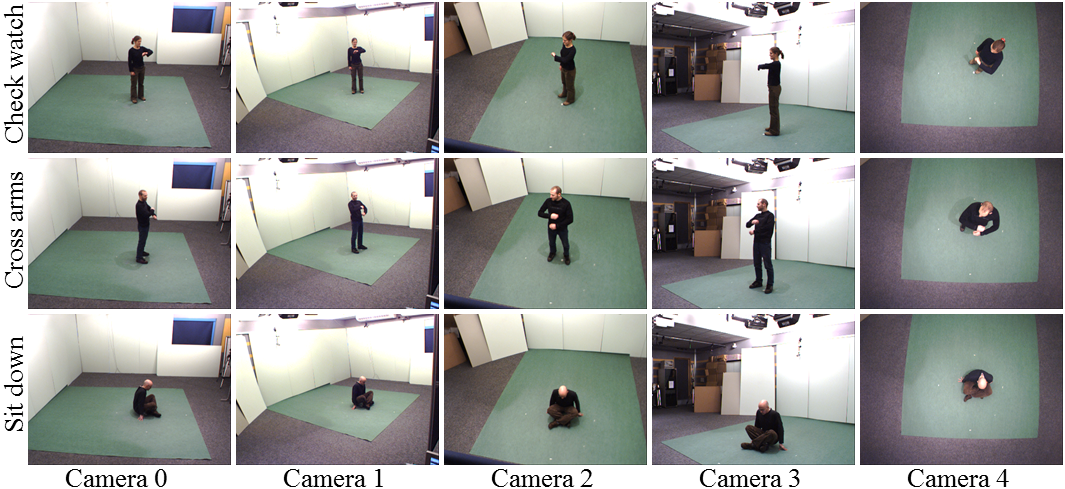}
\caption{\small Sample frames from the IXMAS~\cite{IXMAS} dataset. Each row shows one action captured simultaneously by $5$ cameras.}
\vspace{-5mm}
\label{fig:IXMASSamples}
\end{figure}

\begin{table*}[t] \small
\centering \caption{ \small Accuracy (\%) comparison with state-of-the-art methods under $20$ combinations of source (training) and target (test) views on the IXMAS~\cite{IXMAS} dataset. Each column corresponds to one source$\textemdash$target view pair. The last column shows the average accuracy. The best result of each pair is shown in bold. AOG~\cite{and-or} cannot be applied to this dataset because the 3D joint positions are not provided. Note that DVV and CVP require samples from the target view which are not required by our method.}
\vspace{-2mm}
\setlength{\tabcolsep}{2pt}
    \begin{tabular}{lccccccccccccccccccccc}
    \toprule
    Source$\textemdash$Target & $0\textemdash1$ & $0\textemdash2$& $0\textemdash3$ & $0\textemdash4$ & $1\textemdash0$ & $1\textemdash2$ & $1\textemdash3$ & $1\textemdash4$ & $2\textemdash0$ & $2\textemdash1$ & $2\textemdash3$ & $2\textemdash4$& $3\textemdash0$ & $3\textemdash1$ & $3\textemdash2$& $3\textemdash4$ & $4\textemdash0$ & $4\textemdash1$& $4\textemdash2$ & $4\textemdash3$ & Mean\\
    \midrule \midrule
    DT~\cite{DTraj1} & 93.9 & 64.2 & 81.8 & 27.6 & 87.6 & 66.4 & 75.2 & 22.4 & 70.0 & 83.0 & 73.9 & 53.3 & 75.5 & 77.0 & 67.0 & 34.8 & 42.1 & 25.8 & 63.3 & 48.8 & 61.7\\
    
    Hankelets~\cite{hankelet} & 83.7 & 59.2 & 57.4 & 33.6 & 84.3 &61.6 & 62.8 & 26.9 & 62.5 & 65.2 & 72.0 & 60.1 & 57.1 & 61.5 & 71.0 & 31.2 & 39.6 & 32.8 & 68.1 & 37.4 & 56.4\\
    DVV~\cite{virtualviews} & 72.4 & 13.3 & 53.0 & 28.8 & 64.9 & 27.9 & 53.6 & 21.8 & 36.4 & 40.6 & 41.8 & 37.3 & 58.2 & 58.5 & 24.2 & 22.4 & 30.6 & 24.9 & 27.9 & 24.6 & 38.2\\    
     CVP~\cite{CVV} &  78.5 & 19.5 & 60.4 & 33.4 & 67.9 & 29.8 & 55.5 & 27.0 & 41.0 & 44.9 & 47.0 & 41.0 & 64.3 & 62.2 & 24.3 & 26.1 & 34.9 & 28.2 & 29.8 & 27.6 & 42.2\\ 
    nCTE~\cite{CTE} & {\bf 94.8} & 69.1 & 83.9 & 39.1 & 90.6 & 79.7 & 79.1 & 30.6 & 72.1 & {\bf 86.1} & 77.3 & 62.7 & 82.4 & 79.7 & 70.9 & 37.9 & 48.8 & 40.9 & 70.3 & 49.4 & 67.4\\  
    LRCN~\cite{LRCN} & 66.7 & 63.6 & 39.4 & 16.7 & 60.6 & 51.5 & 36.4 & 16.7 & 63.3 & 27.3 & 50.0 & 30.3 & 45.5 & 47.9 & 42.1 & 15.2 & 14.8 & 13.6 & 18.2 & 13.9 & 36.7 \\
    Action Tube\cite{AT} & 68.5 & 65.2 & 24.2 & 17.0 & 65.8 & 57.6 & 45.5 & 13.3 & 63.6 & 32.7 & 57.0 & 26.1 & 44.2 &  35.5 & 63.9 & 14.5 & 17.0 & 14.8 & 22.1 & 12.7 & 38.1\\
    \midrule      
   NKTM & 92.7 & {\bf 84.2} & {\bf 83.9} & 44.2 & {\bf 95.5} & 77.6 & 86.1 & 40.9 & 82.4 & 79.4 & {\bf 85.8} & 71.5 & 82.4 & 80.9 & 82.7 & {\bf 44.2} & {\bf 57.1} & 48.5 & 78.8 & 51.2 & 72.5\\  
   
   R-NKTM & 92.7 & 80.3 & {\bf 83.9} & {\bf 55.2} & {\bf 95.5} & {\bf 80.6} & {\bf 86.4} & {\bf 47.0} & {\bf 82.7} & 83.6 & 83.6 & {\bf 75.5} & {\bf 85.8} & {\bf 85.2} & {\bf 84.9} & {\bf 44.2} & 56.0 & {\bf 53.0} & {\bf 79.0} & {\bf 52.4} & {\bf 74.1}\\  
    \bottomrule
    \end{tabular}
    \vspace{-4mm}
  \label{tab:IXMAS}
\end{table*}

\vspace{-3mm}\noindent {\bf Dense Trajectories Extraction:}  To generate synthetic dense trajectory descriptors from multiple viewpoints, we use the CMU Motion Capture dataset~\cite{mocap} which contains over $2600$ mocap sequences of different subjects performing a variety of daily-life actions. We remove the short sequences containing less than $15$ frames since dense trajectories require $L=15$ minimum frames. The remaining $2488$ mocap sequences are used for generating synthetic training data to learn the R-NKTM. Each sequence is treated as a different action and given a unique dummy label. We can generate as many different views from the 3D videos as we desire. Using azimuthal angle $\phi \in \Phi=\{0^{\degree}:20^{\degree}:340^{\degree}\}$, and zenith angle $\theta \in \Theta=\{0^{\degree},10^{\degree},30^{\degree},50^{\degree},70^{\degree},90^{\degree}\}$, we generate ($n=108$) camera viewpoints and project the 3D videos. Dense trajectories are then extracted from the 2D projections and clustered into $k=2000$ clusters using $k$-means to make the general codebook. From real videos, we extract dense trajectories using the method by Wang et al.~\cite{DTraj1}. We take the length of each trajectory $L=15$ for both mocap and video sequences. As recommended by~\cite{DTraj1}, we use $8$ spatial scales spaced by a factor of $1/{\sqrt{2}}$ and the dense sampling step size $5$ for video samples. \\

\vspace{-3mm}\noindent {\bf R-NKTM Configuration:} We used multi-resolution search~\cite{ovfit1} to find optimal hyper-parameter values such as weight decay, sparsity and units per layer. The idea is to test some values from a larger parameter range, select a few best configurations and then test again with smaller steps around these values. To optimize the number of R-NKTM layers, we tested networks with increasing number of layers~\cite{Larochelle} and stopped where the performance peaked on our validation data. We used a momentum of $0.9$, weight decay $\lambda_{w}=0.0005$, sparsity parameter $\lambda_{s}=0.5$, and sparsity target $\rho=0.05$.

\vspace{-2mm}
\subsection{IXMAS Dataset}
\label{IXMAS dataset}
This dataset~\cite{IXMAS} consists of synchronized videos observed from $5$ different views including four side views and a top view. It contains $11$ daily-life actions including {\it check watch}, {\it cross arms}, {\it scratch head}, {\it sit down}, {\it get up}, {\it turn around}, {\it walk}, {\it wave}, {\it punch}, {\it kick}, and {\it pick up}. Each action was performed three times by $10$ subjects. Figure~\ref{fig:IXMASSamples} shows examples from this dataset.

We follow the same evaluation protocol as in~\cite{virtualviews,hankelet,CTE} and verify our algorithm on all possible pairwise view combinations. In each experiment, we use all videos from one camera as training samples and then evaluate the recognition accuracy on the video samples from the $4$ remaining cameras. Comparison of the recognition accuracy for $20$ possible combinations of training and test cameras is shown in Table~\ref{tab:IXMAS}.

R-NKTM achieves better recognition accuracy than the NKTM \cite{NKTM} which requires to define a same canonical view for all actions. Moreover, the proposed R-NKTM outperforms the state-of-the-art methods on most view pairs and achieves $74.1\%$ average recognition accuracy which is about $7\%$ higher than the nearest competitor nCTE~\cite{CTE}. It is interesting to note that our R-NKTM can perform much better (about $10\%$ on average) than the nearest competitor nCTE~\cite{CTE} when camera $4$ is considered as either source or target view (see Table~\ref{tab:OIXMAS}). As shown in Fig.~\ref{fig:IXMASSamples}, camera $4$ captured videos from the top view, so the appearance of these videos is completely different from the videos captured from the side views (i.e.~camera $0$ to $3$). Hence, we believe that the recognition results on camera $4$ are the most important for evaluating cross-view action recognition. Moreover, some actions such as {\it check watch, cross arms}, and {\it scratch head} are not available in the mocap dataset. However, our R-NKTM achieves $66.7\%$ average accuracy on these three actions which is about $11\%$ higher than nCTE~\cite{CTE}. This demonstrates that the proposed R-NKTM is able to transfer knowledge across views without requiring all action classes in the learning phase. 

Among the knowledge transfer based methods, DVV~\cite{virtualviews} and CVP~\cite{CVV} did not perform well. The deep learning based methods such as LRCN~\cite{LRCN} and Action Tube~\cite{AT} achieve low accuracy because they were originally proposed for action recognition from a common viewpoint. DT~\cite{DTraj1} achieves a high overall recognition accuracy because the motion trajectories of action videos captured from the side views are similar. However, its average accuracy when camera $4$ is considered as either source or target view, is over $18\%$ lower than our proposed method.

\begin{figure}[t]
\centering
\includegraphics[width=8.3 cm]{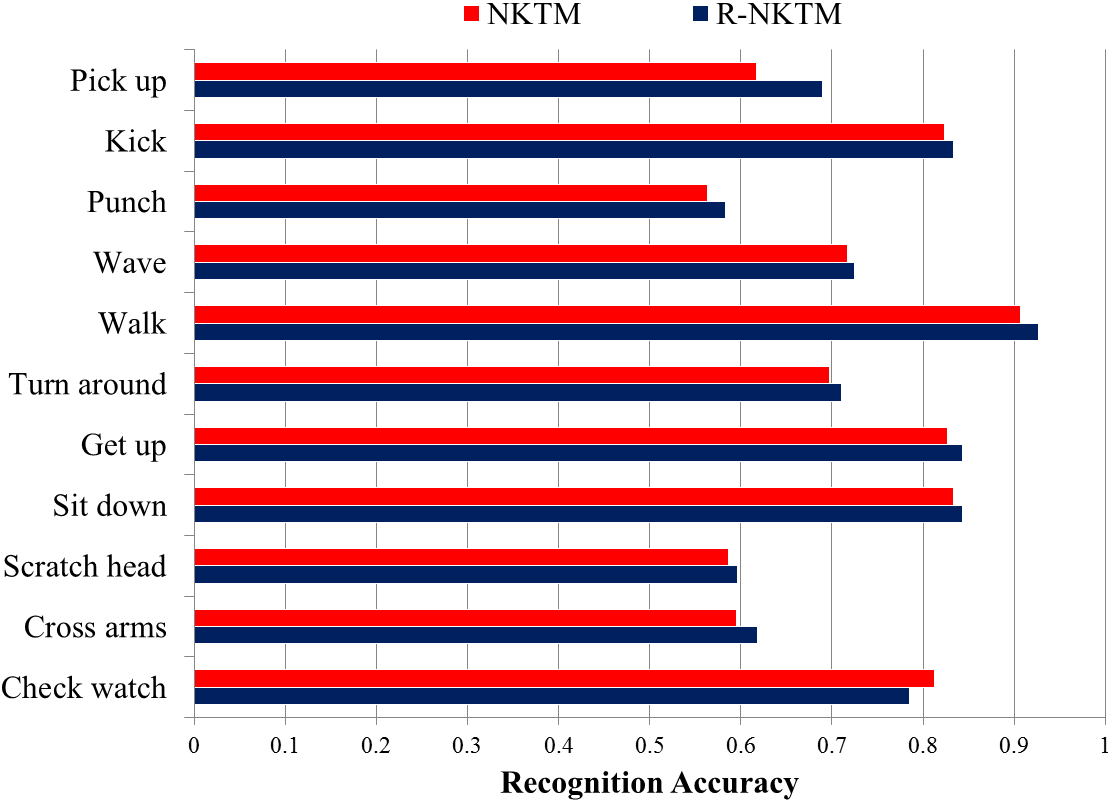}
\vspace{-2mm}
\caption{\small Per class recognition accuracy of our proposed R-NKTM and NKTM~\cite{NKTM} on the IXMAS~\cite{IXMAS} dataset.}
\label{fig:IXMAS-CM}
\end{figure}

Figure~\ref{fig:IXMAS-CM} compares the class specific action recognition accuracies of our proposed R-NKTM with NKTM~\cite{NKTM}. R-NKTM achieves higher accuracies for all action classes excluding {\it check watch}. This demonstrates the effectiveness of our new architecture for cross-view action recognition.

\vspace{-2mm}
\subsection{UWA3D Multiview Activity II Dataset}
\label{UWA3DII dataset}
\begin{figure}[t]
\centering
\vspace{-2mm}
\includegraphics[width=8.3 cm]{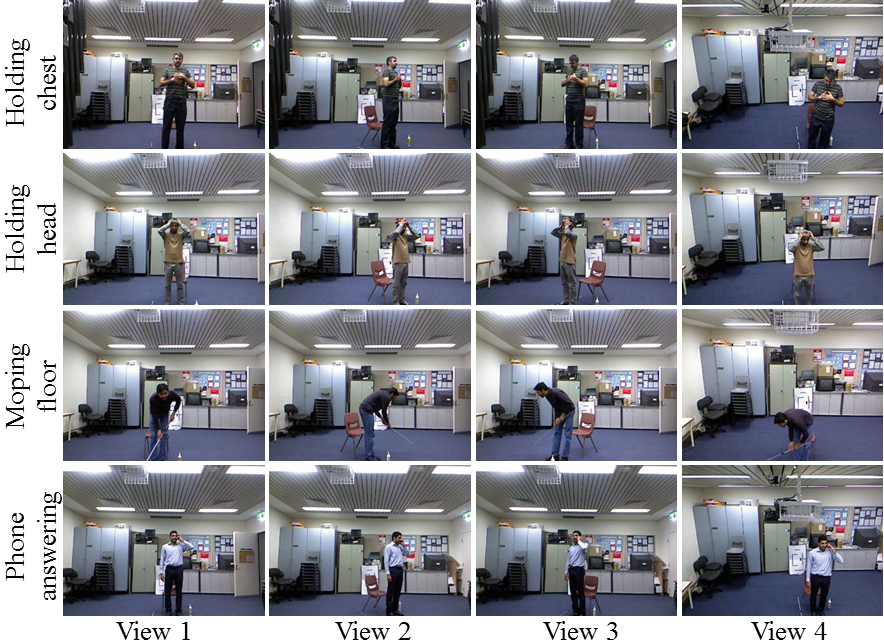}
\vspace{-2mm}
\caption{\small Sample frames from the UWA3D Multiview ActivityII~\cite{MyPAMI} dataset. Each row shows one action acquired from $4$ different views.}
\vspace{-6mm}
\label{fig:UWA3DIISamples}
\end{figure}

This dataset~\cite{MyPAMI} consists of a variety of daily-life human actions performed by $10$ subjects with different scales. It includes $30$ action classes: {\it one hand waving}, {\it one hand Punching}, {\it two hand waving}, {\it two hand punching}, {\it sitting down}, {\it standing up}, {\it vibrating}, {\it falling down}, {\it holding chest}, {\it holding head}, {\it holding back}, {\it walking}, {\it irregular walking}, {\it lying down}, {\it turning around}, {\it drinking}, {\it phone answering}, {\it bending}, {\it jumping jack}, {\it running}, {\it picking up}, {\it putting down}, {\it kicking}, {\it jumping}, {\it dancing}, {\it moping floor}, {\it sneezing}, {\it sitting down (chair)}, {\it squatting}, and {\it coughing}. Each subject performed $30$ actions $4$ times. Each time the action was captured from a different viewpoint (front, top, left and right side views). Video acquisition from multiple views was not synchronous thus there are variations in the actions besides viewpoints. This dataset is challenging because of varying viewpoints, self-occlusion and high similarity among actions. For instance, action {\it drinking} and {\it phone answering} have very similar motion, but the location of hand in these two actions is slightly different. Also, actions like {\it holding head} and {\it holding back} have self-occlusion. Moreover, in the top view, the lower part of the body was not properly captured because of occlusion. Figure~\ref{fig:UWA3DIISamples} shows four sample actions observed from $4$ viewpoints.

\begin{table}[t] \small
\vspace{-2mm}
\centering \caption{\small Average accuracies (\%) on the IXMAS~\cite{IXMAS} dataset e.g.~$C_0$ is the average accuracy when camera 0 is used for training or testing. Each time, only one camera view is used for training and testing. R-NKTM gives the maximum improvement for the most challenging case, Camera 4 (top view).}
\vspace{-2mm}
\setlength{\tabcolsep}{7.5pt} 
    \begin{tabular}{lcccccccccccccccccccccccccccccc}
    \toprule
    Method & $C_0$ & $C_1$ & $C_2$ & $C_3$ & $C_4$ \\
    \midrule \midrule
    DT~\cite{DTraj1} & 67.8 & 66.4 & 67.6 & 66.8 & 39.8 \\  
    Hankelets~\cite{hankelet} & 59.7 & 59.9 & 65.0 & 56.3 & 41.2 \\
    DVV~\cite{virtualviews} &  44.7 & 45.6 & 31.2 & 42.0 & 27.3\\    
    CVP~\cite{CVV} & 50.0 & 49.3 & 34.7 & 45.9 & 31.0\\ 
    nCTE~\cite{CTE} & 72.6 & 72.7 & 73.5 & 70.1 & 47.5\\    
    LRCN~\cite{LRCN} & 46.3 & 40.1 & 43.3 & 36.3 & 17.4\\  
    Action Tube\cite{AT} & 45.7 & 41.7 & 48.5 & 37.2 & 17.2\\  
    \midrule      
    NKTM & 77.8 & 75.2 & 80.3 & 74.7 & 54.6\\ 
    R-NKTM & {\bf 78.4} & {\bf 78.0} & {\bf 80.7} & {\bf 75.8} & {\bf 57.8}\\ 
    \bottomrule
    \end{tabular}
  \label{tab:OIXMAS}
  \vspace{-3mm}
\end{table}

\begin{table*}[t] \small
\centering \caption{\small Comparison of action recognition accuracy (\%) on the UWA3D Multiview ActivityII dataset. Each time two views are used for training and the remaining ones are individually used for testing. Our method achieves the best performance in all cases.}
\vspace{-2mm}
\setlength{\tabcolsep}{2pt}
    \begin{tabular}{lccccccccccccccccccc}
    \toprule
    Sources$\textemdash$Target & $\{1,2\}\textemdash3$ & $\{1,2\}\textemdash4$ & $\{1,3\}\textemdash2$ & $\{1,3\}\textemdash4$ & $\{1,4\}\textemdash2$ & $\{1,4\}\textemdash3$ & $\{2,3\}\textemdash1$ & $\{2,3\}\textemdash4$ & $\{2,4\}\textemdash1$ & $\{2,4\}\textemdash3$ & $\{3,4\}\textemdash1$ & $\{3,4\}\textemdash2$ & Mean\\
    \midrule \midrule
     
     DT~\cite{DTraj1}& 57.1 & 59.9 & 54.1 & 60.6 & 61.2 & 60.8 & 71 & 59.5 & 68.4 & 51.1 & 69.5 & 51.5 & 60.4 \\
     
     Hankelets~\cite{hankelet}& 46.0 & 51.5 & 50.2 & 59.8 & 41.9 & 48.1 & 66.6 & 51.3 & 61.3 & 38.4 & 57.8 & 48.9 & 51.8 \\

     DVV~\cite{virtualviews}& 35.4 & 33.1 & 30.3 & 40.0 & 31.7 & 30.9 & 30.0 & 36.2 & 31.1 & 32.5 & 40.6 & 32.0 & 33.7 \\

     CVP~\cite{CVV}& 36.0 & 34.7 & 35.0 & 43.5 & 33.9 & 35.2 & 40.4 & 36.3 & 36.3 & 38.0 & 40.6 & 37.7 & 37.3 \\
     
     nCTE~\cite{CTE}& 55.6 & 60.6 & 56.7 & 62.5 & 61.9 & 60.4 & 69.9 & 56.1 & 70.3 & 54.9 & 71.7 & 54.1 & 61.2 \\
     
     
     LRCN~\cite{LRCN} & 53.9 & 20.6 & 43.6 & 18.6 & 37.2 & 43.6 & 56.0 & 20.0 & 50.5 & 44.8 & 53.3 & 41.6 & 40.3 \\
     
     Action Tube\cite{AT} & 49.1 & 18.2 & 39.6 & 17.8 & 35.1 & 39.0 & 52.0 & 15.2 & 47.2 & 44.6 & 49.1 & 36.9 & 37.0 \\
     
%
%
     
     \midrule
    
    NKTM & 60.1 & 61.3 & 57.1 & 65.1 & 61.6 & 66.8 & 70.6 & 59.5 & 73.2 & 59.3 & 72.5 & 54.5 & 63.5 \\
    
    R-NKTM & {\bf 64.9} & {\bf 67.7} & {\bf 61.2} & {\bf 68.4} & {\bf 64.9} & {\bf 70.1} & {\bf 73.6} & {\bf 66.5} & {\bf 73.6} & {\bf 60.8} & {\bf 75.5} & {\bf 61.2} & {\bf 67.4} \\
    
    \bottomrule
    \end{tabular}
  \label{tab:UWA3DII}
\end{table*}

We follow~\cite{MyPAMI} and use the samples from two views as training data, and the samples from the remaining views as test data. Table~\ref{tab:UWA3DII} summarizes our results. The proposed R-NKTM significantly outperforms NKTM~\cite{NKTM} and the state-of-the-art methods on all view pairs. The overall accuracy of the view knowledge transfer based methods such as DVV~\cite{virtualviews} and CVP~\cite{CVV} is low because motion and appearance of many actions look very similar across view changes. 

It is interesting to note that our method achieves $67.5\%$ average recognition accuracy which is about $8\%$ higher than than the nearest competitor nCTE~\cite{CTE} when view $4$ is considered as the test view. As shown in Fig.~\ref{fig:UWA3DIISamples}, view $4$ is the top view which is challenging because the lower part of the subject's body was not fully captured by the camera. 

\begin{figure*}[t]
\centering
\includegraphics[width=16 cm]{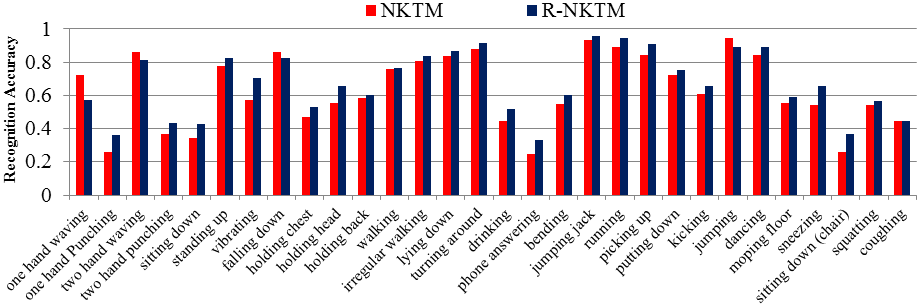}
\vspace{-2mm}
\caption{\small Per class recognition accuracy of the proposed R-NKTM and NKTM~\cite{NKTM} on the UWA3D Multiview ActivityII~\cite{MyPAMI} dataset.}
\vspace{-3mm}
\label{fig:UWA3D-CM}
\end{figure*}

Figure~\ref{fig:UWA3D-CM} compares the class specific action recognition accuracies of R-NKTM and NKTM~\cite{NKTM}. The proposed R-NKTM achieves better recognition accuracy on most action classes. The easiest action to identify is {\it jumping jack} with an average accuracy of $95.4\%$ and the hardest is {\it phone answering} with an average accuracy of $33.3\%$. These results are not surprising, since {\it jumping jack} is one of the activities with the most discriminative trajectories while {\it phone answering} is confused with {\it drinking} because the motion of these actions is very similar.

It is important to note that for many actions in the UWA3D Multiview ActivityII dataset such as {\it holding chest, holding head, holding back, sneezing} and {\it coughing}, there are no similar actions in the CMU mocap dataset. However, our method still achieves high recognition accuracies for these actions. This demonstrates the effectiveness and generalization ability of our proposed model for representing human actions from unseen and unknown views in a view-invariant space.

\vspace{-2mm}
\subsection{N-UCLA Multiview Action3D Dataset}
\label{N-UCLA dataset}
\begin{figure}[t]
\centering
\includegraphics[width=8.3 cm]{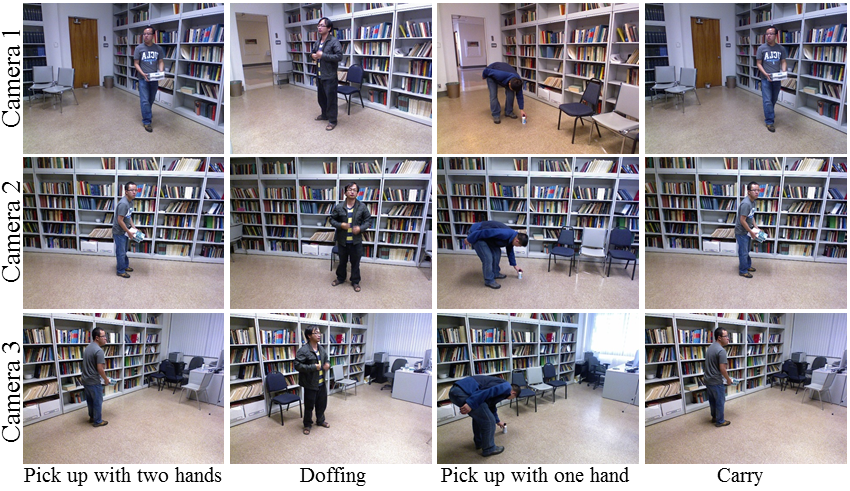}
\vspace{-1mm}
\caption{\small Sample frames from Northwestern-UCLA Multiview Action3D dataset~\cite{and-or}. Each column shows a different action.}
\vspace{-5mm}
\label{fig:N-UCLASamples}
\end{figure}

This dataset~\cite{and-or} contains RGB, depth and skeleton data captured simultaneously by $3$ Kinect cameras. The dataset consists of $10$ action categories including {\it pick up with one hand}, {\it pick up with two hands}, {\it drop trash}, {\it walk around}, {\it sit down}, {\it stand up}, {\it donning}, {\it doffing}, {\it throw}, and {\it carry}. Each action was performed by $10$ subjects from $1$ to $6$ times. Fig.~\ref{fig:N-UCLASamples} shows some examples. This dataset is very challenging because the subjects performed some {\it walking} within most actions and the motion of some actions such as {\it carry} and {\it walk around} are very similar. Moreover, most activities involve human-object interactions.

We follow \cite{and-or} and use the samples from the first two cameras for training and samples from the remaining camera for testing. The comparison of the recognition accuracy is shown in Table~\ref{tab:N-UCLA}. The proposed R-NKTM again outperforms the NKTM~\cite{NKTM} and achieves the highest recognition accuracy. 

\begin{figure}
\centering
\includegraphics[width=8.3 cm]{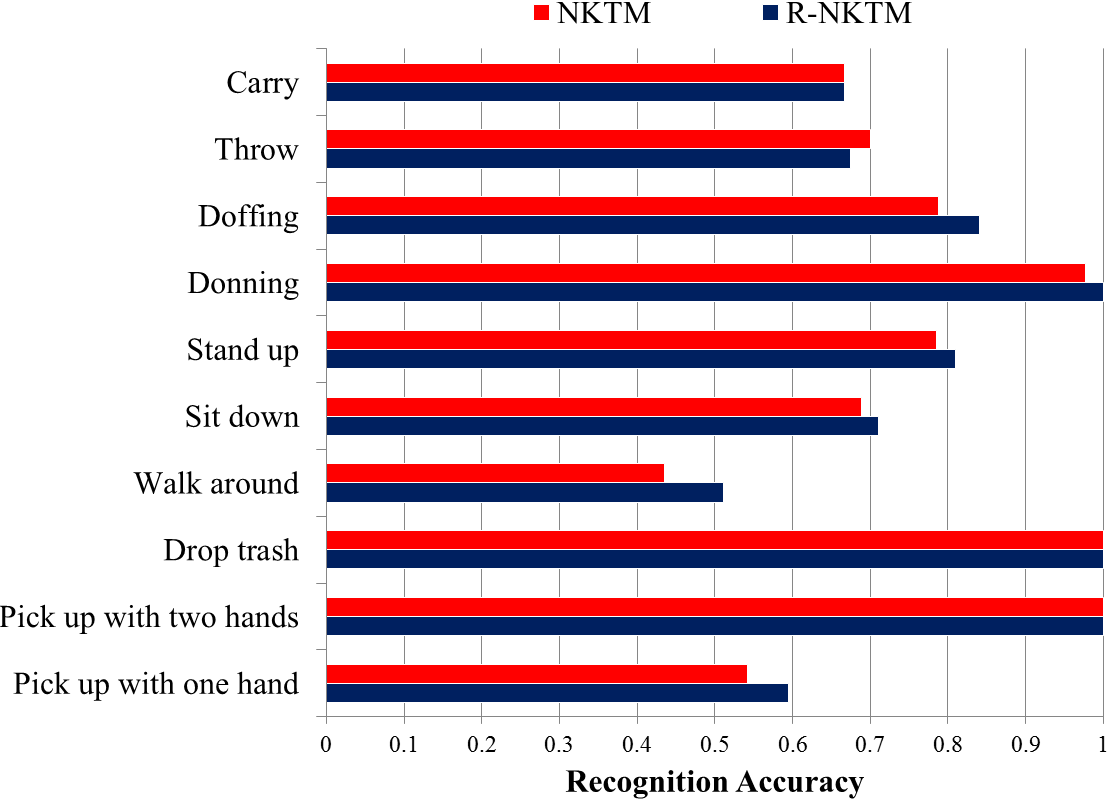}
\vspace{-2mm}
\caption{\small Per class recognition accuracy of the proposed R-NKTM and NKTM~\cite{NKTM} on the Northwestern-UCLA Action3D dataset~\cite{and-or}.}
\vspace{-1mm}
\label{fig:N-UCLA-CM}
\end{figure}

Figure~\ref{fig:N-UCLA-CM} compares the per action class recognition accuracy of our proposed R-NKTM and NKTM~\cite{NKTM}. Our method achieves higher accuracy than NKTM~\cite{NKTM} for most action classes. Note that a search for some actions such as {\it donning, doffing} and {\it drop trash} returns no results on the CMU mocap dataset~\cite{mocap} used to learn our R-NKTM. However, our method still achieves $76.8\%$ average recognition accuracy on these three actions which is about $10\%$ higher than nCTE~\cite{CTE}. Moreover, {\it walk around} and {\it carry} have maximum confusion with each other because the motion of these actions are very similar.\\

\begin{table} \small
\centering \caption{\small Accuracy (\%) on the N-UCLA Multiview dataset~\cite{and-or} when the samples from the first two cameras are used for training and the samples from the third camera for testing. DVV and CVP use samples from the target view. AOG requires the joint positions of training samples. Our method neither requires target view samples nor joint positions.}
\vspace{-2mm}
    \begin{tabular}{lcccccccccccccccccccccccccccccc}
    \toprule
    Method & Accuracy \\
    \midrule \midrule
    DT~\cite{DTraj1} & 72.7 \\
    Hankelets~\cite{hankelet} & 45.2 \\
    DVV~\cite{virtualviews} & 58.5 \\ 
    CVP~\cite{CVV} & 60.6 \\     
    nCTE~\cite{CTE}  & 68.6 \\
    AOG~\cite{and-or} & 73.3 \\
    LRCN~\cite{LRCN} & 64.7 \\
    Action Tube\cite{AT} & 61.5 \\
    \midrule      
    NKTM & 75.8 \\  
    R-NKTM & {\bf 78.1}\\  
    \bottomrule
    \end{tabular}
  \label{tab:N-UCLA}
\end{table}

\vspace{-4mm}
\subsection{UCF Sports Dataset}
\label{N-UCLA dataset}
While the focus of the proposed approach is on action recognition from unknown and unseen views, we also evaluate its performance for recognizing actions from previously seen views to have a baseline and to show that our method performs equally good when the viewpoint of the test action is not novel. The evaluation is performed on the UCF Sports dataset~\cite{UCFSports} containing videos from sports broadcasts in a wide range of scenes. As recommended in~\cite{UCFSports}, we use the Leave-One-Out (LOO) cross-validation scheme. We compare our proposed method to the dense trajectory based method (DT)~\cite{DTraj1}. We choose DT~\cite{DTraj1} as our baseline because it is most relevant to our work as it employs dense trajectory descriptors. As shown in Table~\ref{tab:UCFSports}, using only trajectory descriptors, our method achieves $1.5\%$ higher accuracy than DT~\cite{DTraj1}. However, combining HOG, HOF, and MBH descriptors with the trajectory descriptor significantly increases the recognition accuracy of DT~\cite{DTraj1} by $13\%$. Similarly, adding these features to our cross-view action descriptor significantly improves the accuracy of our method to $90\%$ which is about $2\%$ higher than DT~\cite{DTraj1}. 

Combining the view dependent HOG, HOF and MBH descriptors with our cross-view descriptor also improves the recognition accuracy for the multiview case especially when the difference between the viewpoints is not large. Table~\ref{tab:IXMAS2} shows comparative results of combined descriptors and the cross-view trajectory only descriptors on the IXMAS dataset. The accuracy of most source$\textemdash$target combinations from side views have improved by using the combined features. This is because the appearance of these views is quite similar.

\begin{table*}[t] \small
\centering \caption{ \small  Effects of combining HOG, HOF, MBH with our proposed cross-view descriptor on the IXMAS~\cite{IXMAS} dataset}
\vspace{-2mm}
\setlength{\tabcolsep}{1.7pt}
    \begin{tabular}{|l|c|c|c|c|c|c|c|c|c|c|c|c|c|c|c|c|c|c|c|c|c|}
    \hline
    Source$\textemdash$Target & $0\textemdash1$ & $0\textemdash2$& $0\textemdash3$ & $0\textemdash4$ & $1\textemdash0$ & $1\textemdash2$ & $1\textemdash3$ & $1\textemdash4$ & $2\textemdash0$ & $2\textemdash1$ & $2\textemdash3$ & $2\textemdash4$& $3\textemdash0$ & $3\textemdash1$ & $3\textemdash2$& $3\textemdash4$ & $4\textemdash0$ & $4\textemdash1$& $4\textemdash2$ & $4\textemdash3$ & Mean\\
    \hline \hline
   
   R-NKTM (Traj. only) & 92.7 & 80.3 & 83.9 & {\bf 55.2} & 95.5 & 80.6 & 86.4 & {\bf 47.0} & {\bf 82.7} & 83.6 & 83.6 & {\bf 75.5} & 85.8 & 85.2 & 84.9 & {\bf 44.2} & {\bf 56.0} & {\bf 53.0} & 79.0 & {\bf 52.4} & {\bf 74.1}\\  
   \hline
   R-NKTM (all) & {\bf 96.7} & {\bf 80.3} & {\bf 89.1} & 51.5 & {\bf 96.7} & {\bf 80.9} & {\bf 88.5} & 43.3 & 79.7 & {\bf 87.9} & {\bf 84.8} & 73.9 & {\bf 86.1} & {\bf 87.9} & {\bf 87.9} & 43.3 & 54.5 & 50.3 & {\bf 84.2} & 52.4 & {\bf 75.0}\\  
    \hline
    \end{tabular}
  \label{tab:IXMAS2}
\end{table*}

\begin{table}[t] \small
\centering \caption{\small Comparison of action recognition accuracy (\%) on the UCF Sports dataset. 
}
\vspace{-2mm}
\setlength{\tabcolsep}{3pt}
    \begin{tabular}{|l|c|c|}
    \hline
    Method &  Only trajectories & HOG+HOF+MBH+Traj.\\
    \hline \hline
    DT~\cite{DTraj1} & 75.2 & 88.2 \\
    \hline      
    R-NKTM & {\bf 76.7} & {\bf 90.0} \\  
    \hline
    \end{tabular}
  \label{tab:UCFSports}
  \vspace{-3mm}
\end{table}

\vspace{-2mm}
\subsection{Effects of Concatenating Virtual Views}
\begin{figure*}[t]
\centering
\includegraphics[width=\linewidth]{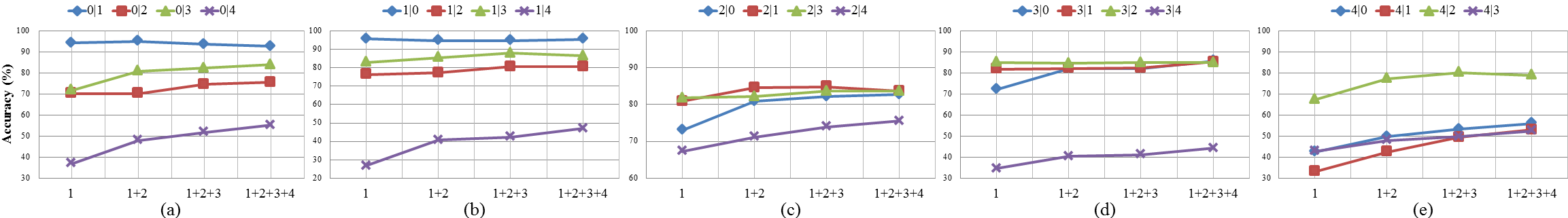}
\vspace{-5mm}
\caption{\small IXMAS dataset: Effects of adding features from different layers to the cross-view action descriptor e.g.~$1+2+3$ means that the descriptor is built by concatenating features from the source view, virtual view 1 and virtual view 2 as shown in Fig.~\ref{fig:TrainingPhase}.}
\vspace{-3mm}
\label{fig:IXMAS_layers_acc}
\end{figure*}

\begin{figure*}[t]
\centering
\includegraphics[width=\linewidth]{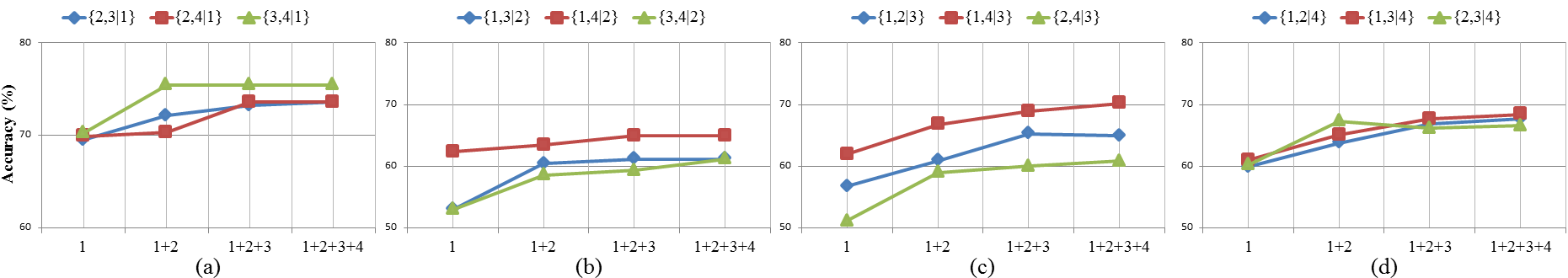}
\vspace{-5mm}
\caption{\small UWA3DII dataset: Effects of adding features from different layers to the cross-view action descriptor}
\vspace{-3mm}
\label{fig:UWA3DII_layers_acc}
\end{figure*}

We evaluate the intermediate performance of our cross-view descriptor by sequentially adding the virtual views. Figures~\ref{fig:IXMAS_layers_acc} and~\ref{fig:UWA3DII_layers_acc} show the recognition accuracy on IXMAS and UWA3DII datasets respectively for all possible source$\textemdash$target view pairs. For most source$\textemdash$target view pairs of IXMAS dataset, the accuracy increases as more virtual views are added to the cross-view action descriptor. The maximum incremental gain is obtained when camera $4$ (top view) is used as training or test view. The minimum gain is for $0\textemdash1$ view pair because the viewpoints of these cameras are very similar. Thus the raw trajectory descriptors already achieve high accuracy. Fig.~\ref{fig:UWA3DII_layers_acc} shows that for all source$\textemdash$target view pairs of UWA3DII dataset, the recognition accuracy increases by adding virtual views to the descriptor.

\vspace{-2mm}
\subsection{Computation Time}
It is interesting to note that our technique outperforms the current cross-view action recognition methods on the IXMAS~\cite{IXMAS}, UWA3DII~\cite{MyPAMI} and N-UCLA~\cite{and-or} datasets by transferring knowledge across views using the same R-NKTM learned without supervision (without real action labels). Therefore, compared to existing cross-view action recognition techniques, the proposed R-NKTM is more general and can be used in on-line action recognition systems. More precisely, the cost of adding a new action class using our approach in an on-line system is equal to SVM training. On the other hand, this situation is computationally expensive for most existing techniques especially for our nearest competitors~\cite{and-or,CTE} as shown in Table~\ref{tab:traincomplexity}. For instance nCTE~\cite{CTE} requires to perform computationally expensive spatio-temporal matching for each video sample of the new action class. Similarly, AOG~\cite{and-or} needs to retrain the AND/OR structure and tune its parameters. Table~\ref{tab:traincomplexity} compares the computational complexity of the proposed method with AOG~\cite{and-or} and nCTE~\cite{CTE}. Compared to AOG~\cite{and-or} and nCTE~\cite{CTE}, the training time of the proposed method for adding a new action class is negligible. Thus, it can be used in an on-line system. Moreover, the test time of the proposed method is much faster than AOG~\cite{and-or} and comparable to nCTE~\cite{CTE}. However, nCTE~\cite{CTE} requires $30$GB memory to store the augmented samples whereas our model requires $57$MB memory to store the learned R-NKTM and the general codebook.

\begin{table}[t] \small
\centering \caption{\small Computation time (in minutes) including feature extraction on the N-UCLA dataset~\cite{and-or} when cameras 1, 2 videos are used as source and camera 3 videos are used as target views. Train+1 is the time required to add a new action class after training with $9$ classes. Testing time is for classifying $429$ action videos.}
\vspace{-2mm}
\setlength{\tabcolsep}{3pt}
    \begin{tabular}{|l|c|c|}
    \hline
    Method &  Train+1 & Testing\\
    \hline \hline
    AOG~\cite{and-or} & 780 & 240 \\
    nCTE~\cite{CTE}  & 19 & 12 \\
    \hline      
    R-NKTM & {\bf 0.52} & {\bf 12} \\  
    \hline
    \end{tabular}
    \vspace{-4mm}
  \label{tab:traincomplexity}
\end{table}

\section{Conclusion}
We presented an algorithm for unsupervised learning of a Robust Non-linear Knowledge Transfer Model (R-NKTM) for cross-view action recognition. We call it unsupervised because the labels used to learn the R-NKTM are just dummy labels and do not correspond to actions that we want to recognize. The proposed R-NKTM is scalable as it needs to be trained only once using synthetic data and generalizes well to real data. We presented a pipeline for generating a large corpus of synthetic training data required for deep learning. The proposed method generates realistic 3D videos by fitting 3D human models to real motion capture data. The 3D videos are projected on 2D plains corresponding to a large number of viewing directions and their dense trajectories are calculated. Using this approach, the dense trajectories are realistic and easy to compute since the correspondence between the 3D human poses is known a priori. A general codebook is learned from these trajectories using k-means and then used to represent the synthetic trajectories for R-NKTM learning as well as the trajectories extracted from real videos during training and testing. The major strength of the proposed R-NKTM is that a single model is learned to transform any action from any viewpoint to its respective high level representation. Moreover, action labels or knowledge of the viewing angles are not required for R-NKTM learning or R-NKTM based representation of real video data. To represent actions in real video sequences, their dense trajectories are coded with the general codebook and forward propagated through the R-NKTM. A simple linear SVM classifier was used to show the strength of our model. Experiments on benchmark multiview datasets show that the proposed approach outperforms existing state-of-the-art.

\section*{Acknowledgment}
\addcontentsline{toc}{section}{Acknowledgment}
This research was supported by ARC Discovery Grants DP110102399 and DP160101458.

\ifCLASSOPTIONcaptionsoff
  \newpage
\fi



%
\bibliographystyle{IEEEtran}
\bibliography{MyTPAMI}

\begin{IEEEbiography}[{\includegraphics[width=1in,height=1.25in,clip,keepaspectratio]{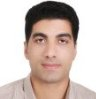}}]
{Hossein Rahmani} received his B.Sc. degree in Computer Software Engineering in 2004 from Isfahan University of Technology (IUT), Isfahan, Iran and the M.S. degree in Software Engineering in 2010 from Shahid Beheshti University (SBU), Tehran, Iran. He has published several papers in conferences and journals such as CVPR, ECCV, WACV, and TPAMI. His research interests include computer vision, 3D shape analysis, and pattern recognition. He is currently working towards his PhD degree in computer science from The University of Western Australia. His current research is focused on RGB-Depth based human activity recognition.
\end{IEEEbiography}

\begin{IEEEbiography}[{\includegraphics[width=1in,height=1.25in,clip,keepaspectratio]{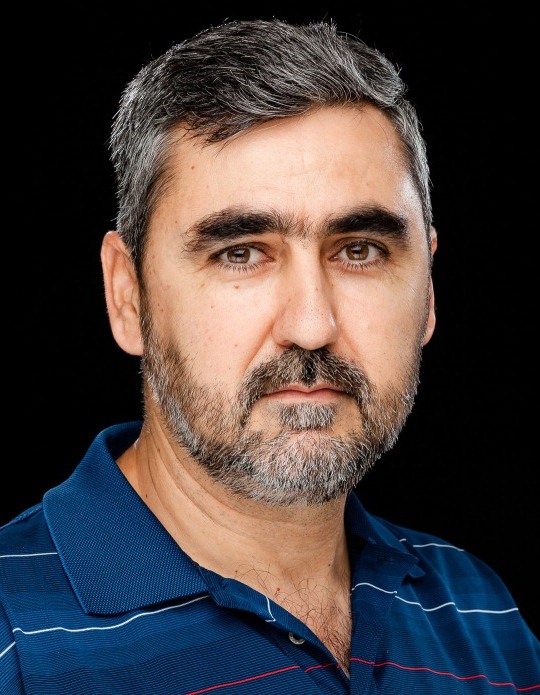}}]
{Ajmal Mian} completed his PhD from The University of Western Australia in 2006 with distinction and received the Australasian Distinguished Doctoral Dissertation Award from Computing Research and Education Association of Australasia. He received two prestigious nationally competitive fellowships namely the Australian Postdoctoral Fellowship in 2008 and the Australian Research Fellowship in 2011. He received the UWA Outstanding Young Investigator Award in 2011, the West Australian Early Career Scientist of the Year award in 2012 and the Vice-Chancellor’s Mid-Career Research Award in 2014. He is currently with the School of Computer Science and Software Engineering, The University of Western Australia. His research interests include computer vision, action recognition, 3D shape analysis, 3D facial morphometrics, hyperspectral image analysis, machine learning and biometrics.
\end{IEEEbiography}

\begin{IEEEbiography}[{\includegraphics[width=1in,height=1.25in,clip,keepaspectratio]{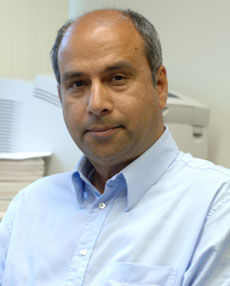}}]
{Mubarak Shah}, the Trustee chair professor of computer science, is the founding director of the Center for Research in Computer Vision at the University of Central Florida (UCF). He is an editor of an international book series on video computing, editor-in-chief of Machine Vision and Applications journal, and an associate editor of ACM Computing Surveys journal. He was the program co-chair of the IEEE Conference on Computer Vision and Pattern Recognition (CVPR) in 2008, an associate editor of the IEEE Transactions on Pattern Analysis and Machine Intelligence, and a guest editor of the special issue of the International Journal of Computer Vision on Video Computing. His research interests include video surveillance, visual tracking, human activity recognition, visual analysis of crowded scenes, video registration, UAV video analysis, and so on. He is an ACM distinguished speaker. He was an IEEE distinguished visitor speaker for 1997-2000 and received the IEEE Outstanding Engineering Educator Award in 1997. He is a fellow of the IEEE, AAAS, IAPR, and SPIE. 
\end{IEEEbiography}

\balance
%

%
%
%
%



\end{document}